\documentclass[a4paper,fleqn]{cas-sc}

\usepackage[numbers]{natbib}

\def\tsc#1{\csdef{#1}{\textsc{\lowercase{#1}}\xspace}}
\tsc{WGM}
\tsc{QE}

\usepackage{amssymb}
\usepackage{times}

\usepackage{algorithm}
\usepackage{algorithmic}
\usepackage{subfigure}
\usepackage{hyperref}
\hypersetup{colorlinks=true,
linkcolor=cyan,citecolor=cyan,urlcolor=cyan
}

\newif\ifmarkrevision
\markrevisionfalse
\newcommand{\revision}[1]{\ifmarkrevision
{\textcolor{blue}{#1}}
\else {#1} \fi}

\begin{document}
\let\WriteBookmarks\relax
\def\floatpagepagefraction{1}
\def\textpagefraction{.001}

\shorttitle{Customizing Graph Neural Networks using Path Reweighting}    

\shortauthors{J. Chen, Y. Wang, M. Zeng et al.}  

\title [mode = title]{Customizing Graph Neural Networks using Path Reweighting}  



%
\author[inst1,inst7]{Jianpeng Chen}[orcid=0000-0001-7599-9961]\fnmark[1]\ead{jianpengc@vt.edu}
\credit{Conceptualization, Methodology, Software, Visualization, Investigation, Writing - Original Draft, Writing - Review \& Editing}
\author[inst2]{Yujing Wang}\fnmark[1]\ead{yujwang@pku.edu.cn}
\credit{Conceptualization, Methodology, Formal analysis, Writing - Original Draft, Writing - Review \& Editing}
\author[inst3]{Ming Zeng}\fnmark[1]\ead{ming.zeng@sv.cmu.edu}
\credit{Conceptualization, Methodology, Formal analysis,  Writing - Review \& Editing}
\author[inst1]{Zongyi Xiang}\ead{xzongyi@foxmail.com}
\credit{Validation, Visualization, Writing - Original Draft}
\author[inst4]{Bitan Hou}\ead{houbitan@sjtu.edu.cn}
\credit{Conceptualization}
\author[inst2]{Yunhai Tong}\ead{yhtong@pku.edu.cn}
\credit{Writing - Review \& Editing}
\author[inst6]{Ole Mengshoel}\ead{ole.j.mengshoel@ntnu.no}
\credit{Writing - Review \& Editing}
\author[inst1,inst5]{Yazhou Ren}\cormark[1]\ead{yazhou.ren@uestc.edu.cn}
\credit{Supervision, Project administration, Funding acquisition, Writing - Review \& Editing}

\address[inst1]{School of Computer Science and Engineering, University of Electronic Science and Technology of China}
\address[inst7]{Department of Computer Science, Virginia Tech}
\address[inst2]{Key Laboratory of Machine Perception, MOE, School of EECS, Peking University}
\address[inst3]{Electrical and Computer Engineering, Carnegie Mellon University}
\address[inst4]{Shanghai Jiao Tong University}
\address[inst5]{Shenzhen Institute for Advanced Study, University of Electronic Science and Technology of China}
\address[inst6]{Department of Computer Science, Norwegian University of Science and Technology}

\cortext[1]{Corresponding author.}
\fntext[1]{Equal contribution.}




\begin{abstract}
Graph Neural Networks (GNNs) have been extensively used for mining graph-structured data with impressive performance. However, because these traditional GNNs do not distinguish among various downstream tasks, embeddings embedded by them are not always effective. Intuitively, paths in a graph imply different semantics for different downstream tasks. Inspired by this, we design a novel GNN solution, namely Customized Graph Neural Network with Path Reweighting (CustomGNN for short). Specifically, the proposed CustomGNN can automatically learn the high-level semantics for specific downstream tasks to highlight semantically relevant paths as well to filter out task-irrelevant noises in a graph. Furthermore, we empirically analyze the semantics learned by CustomGNN and demonstrate its ability to avoid the three inherent problems in traditional GNNs, i.e., over-smoothing, poor robustness, and overfitting. In experiments with the node classification task, CustomGNN achieves state-of-the-art accuracies on three standard graph datasets and four large graph datasets. The source code of the proposed CustomGNN is available at \url{https://github.com/cjpcool/CustomGNN}.
\end{abstract}




\maketitle

\section{Introduction}
\label{Introduction}
\begin{figure}
\centering
\subfigure[Example of node classification task.]{
    \includegraphics[width=0.2\linewidth]{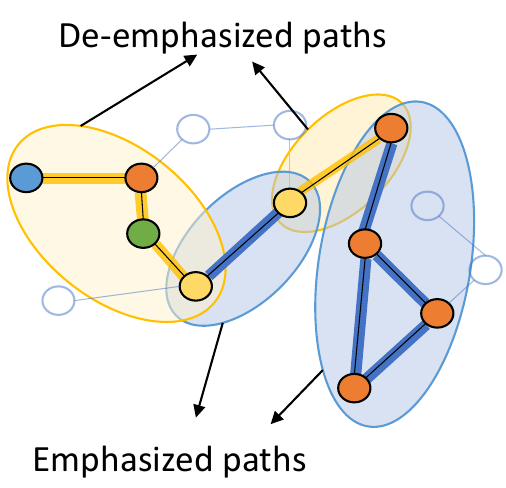}
    \label{Example:cls}
    }
\subfigure[Example of medical Q\&A task on Wikipedia knowledge graph.]{
    \includegraphics[width=0.5\linewidth]{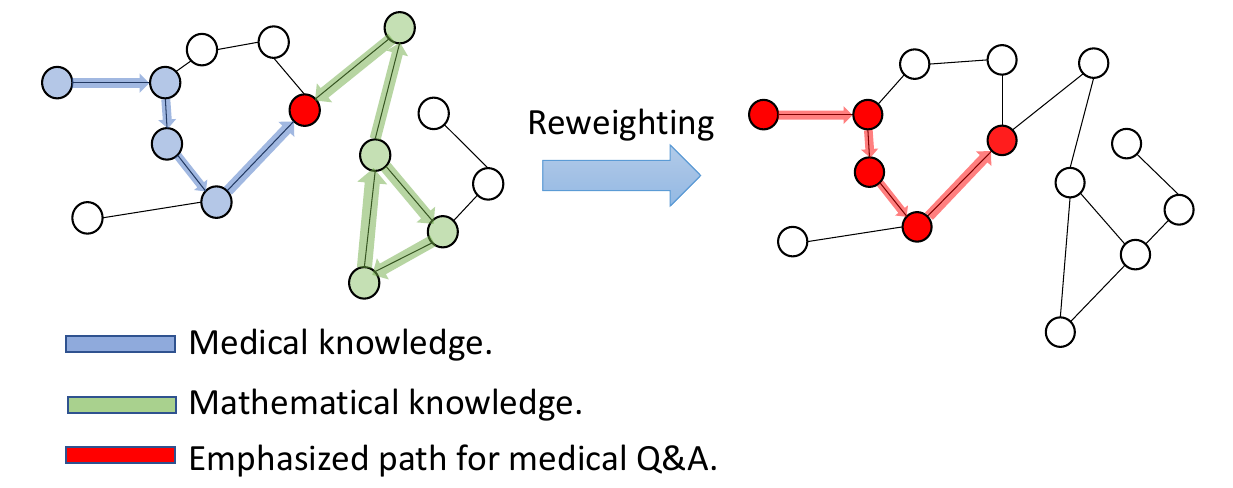}
    \label{Example:KG}
    }
\subfigure[Traditional GNNs can not distinguish paths.]{
    \includegraphics[width=0.2\linewidth]{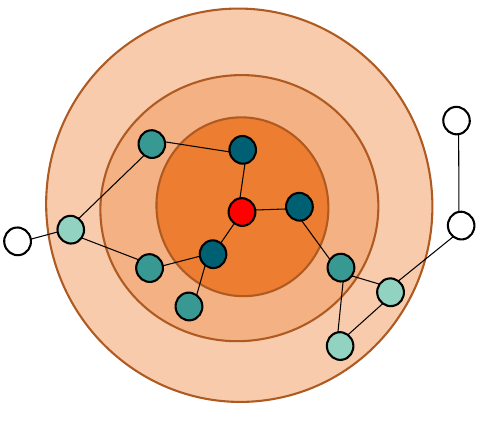}
    \label{Example:gnnIssues}
    }
\caption{Examples to illustrate the intuition of path re-weighting. (a) For node classification, the paths with intra-class nodes should be emphasized (blue paths) while paths with inter-class nodes (orange paths) should be de-emphasized. (b) For medical Q\&A task, each path is a piece of knowledge. Only the Q\&A task-relevant paths should be emphasized. (c) Traditional GNNs aggregate neighbors order by order, neglecting the difference between paths.}
\end{figure}

\revision{Graph-structured datasets are receiving increasing attention since they reflect real-world data such as biological networks, social networks, citation networks, and World Wide Web. The wide range of applications demonstrates the value of mining graph structures in addressing a multitude of practical issues. Many works focus on semi-supervised learning with graph data~\cite{2018Ding,GENG2022126,GCN,xu2021graphsad,2016Revisiting}; they model the non-Euclidean space of a graph and learn structural information. Among these works, the most notable branch of studies is graph neural networks (GNNs), which embed graph-structured data through feature propagation~\cite{chen2023shared,chen2020simple,hamilton2017inductive,HUANG2022286,GCN,LI202250,GAT,DBLP:conf/icml/WuSZFYW19}.
}
Intuitively, given a graph, different applications require different path weighting strategies to emphasize task-related paths and de-emphasize irrelevant ones. As shown in Figure~\ref{Example:cls}, if we aim to optimize a node classification model by exploiting a paper citation graph, the paths between nodes from intra-class should be emphasized and the paths between nodes from inter-class should be de-emphasized. \revision{Concretely, given a citation graph where the relevance among categories is captured by path weight, if the downstream task involves classifying all papers, greater emphasis will be placed on paths that contain more nodes in the same category.}
If the downstream task is to find papers in a specific category (e.g., AI category), then, only the paths containing more nodes relevant to this category (AI) will be selected. Indeed, this motivation is experimentally demonstrated by our analysis (see Sections~\ref{sec:SemanticExp} and \ref{sec:weightCorrelation}).
Another example is about medical Q\&A system on knowledge graph (shown in Figure~\ref{Example:KG}): If we tend to enhance the performance of medical Q\&A by exploiting Wikipedia knowledge graph, some general knowledge that is not related to the medical domain (e.g., mathematical knowledge) should be de-emphasized.

However, traditional GNNs ignore the differences among paths as shown in Fiture~\ref{Example:gnnIssues}, which leads to three potential problems. 1) \emph{Over-smoothing}: \citet{2017neural} states that GNNs can be regarded as a message-passing method with a low-passing filter, resulting in exponentially information lost with the increment of propagation steps, i.e., the problem of over-smoothing~\cite{nt2019revisiting,oono2021graph}. Although some current GNNs, such as GRAND~\cite{NEURIPS2020_fb4c835f}, GraphMix~\cite{2019GraphMix} and S\(^2\)GNN~\cite{zhu2021simple}, have considered multi-hop information, they are still based on the theory of message passing and ignore the differences among different paths, which also leads to the injection of task-irrelevant information and the loss of important information in the message-passing process. \revision{According to these works, the differentiation of different paths with varying hops through the message-passing method appears to be challenging.}
\revision{However, if we consider it from a different perspective, by extracting the high-level semantics of various paths and directly aggregating the high-order information instead of passing messages hop by hop, we can prevent information loss and noise injection while distinguishing between different paths.}
2) \emph{Poor robustness:} On the other hand, the poor robustness to graph attack~\cite{WEI2023166,2019attckGCN,2018attack} of traditional GNNs also comes from that they cannot filter out irrelevant paths and noises (e.g., the paths containing various categories in Figure~\ref{Example:cls} and the paths irrelated to medical knowledges in Figure~\ref{Example:KG}).
\revision{Essentially, the propagation step aggregates all neighborhoods' information with noises, while some noisy nodes and edges may dominate the representation learning procedure. Fortunately, this issue has the chance to be addressed by de-emphasizing the paths containing these noisy nodes and edges.}
3) \emph{Overfitting}: \revision{The third issue arising from semi-supervised learning is that traditional training processes for GNNs can easily overfit the limited labeled data~\cite{2009Semi}. To address this problem, we propose a multi-perspective regularization approach and introduce a novel triplet loss, which not only enhances semantic understanding but also mitigates the risk of overfitting.}

Considering these aspects, we propose Customized Graph Neural Network with Path Reweighting (CustomGNN). CustomGNN differentiates paths by extracting their high-level semantics and customizes node (graph) embeddings by aggregating high-order information directly for specific downstream tasks. In detail, CustomGNN first generates multi-perspective subgraphs. Then, a neural network with LSTM-based~\cite{LSTM} path weighting strategy is proposed, which calculates multiple path attention maps by extracting specific high-level semantic information from these subgraphs. After that, CustomGNN aggregates features through these path attention maps so as to derive the task-oriented embeddings. Finally, we fuse these task-oriented embeddings with the generic embeddings from our multi-hop GNN. Consequently, CustomGNN can automatically filter out noise and encode the most useful structural information for specific tasks.

Furthermore, we analyze the semantics of path weights for the node classification task, which demonstrates the intuition that CustomGNN can emphasize the task-related paths while de-emphasize the irrelevant ones. We also empirically prove that our method can effectively avoid the issues of over-smoothing and non-robustness, while mitigating the issue of overfitting. Moreover, we show that the novel components proposed in this paper bring significant improvement in ablation experiments.
Our contributions are concluded as follows.
\begin{itemize}
    \item We experimentally analyzed the relevance between path weights and semantics, based on which a novel GNN architecture is proposed, named CustomGNN. CustomGNN extracts task-oriented semantic features from re-weighted paths in subgraphs and infuses the extracted information with a standard multi-hop GNN.
    \revision{With this design, the proposed design of CustomGNN enables automatic noise filtering and encoding of the most relevant path structural information tailored to a specific task.}
    \item We propose two unsupervised loss functions to regularize the learning procedure of CustomGNN. The first regularization involves utilizing random walk to generate multi-perspective subgraphs and incorporating consistency loss in-between. In the second regularization term, unlabeled data is utilized to produce pseudo-labels for enhancing semantics, followed by the application of triplet loss based on these pseudo-labels. This semantic-based design provides a new solution to adopt triplet loss with pseudo-labels for graph learning.
    \item As demonstrated by comprehensive experiments and analyses, the proposed CustomGNN model achieves state-of-the-art (SOTA) results on seven widely-used graph node classification datasets. It also shows high interpretability and effectively mitigates the issues of over-smoothing, non-robustness, and overfitting.
\end{itemize}
\section{Related Work}
\label{Related}
\subsection{Graph Neural Networks} 
\revision{The aim of GNNs is to learn the information structured in a graph. They achieve this by aggregating the information from neighboring nodes and learning a vector representation for each node. The core difference among different GNNs lies in their method of message passing (aggregation)~\cite{2017neural}.} For example, to aggregate information, GCN~\cite{GCN} proposed a graph convolution layer. GAT~\cite{GAT} presented graph attention layer, utilizing the feature similarity among nodes. Following GAT, \citet{GENG2022126} introduced graph correlated attention recurrent neural network for the task of multivariate time series forecasting. \citet{he2022neighbors} proposed graph selective attention networks (SAT) to filter out irrelevant neighbors when computing attention. In contrast to GAT, \citet{Zhang2020Adaptive} and \citet{he2021learning} further discussed the significance of structural information when computing attention. Specifically, \citet{Zhang2020Adaptive} proposed adaptive structural fingerprint (ADSF) which perceives structures by Gaussian decay and nonparametric decay; \citet{he2021learning} introduced graph conjoint attention networks (CATs) where the structural information is fused by a proposed structural intervention generation module. \revision{However, the aforementioned two attention-based methods only attempt to perceive structural and feature information, which can be considered a subset of what CustomGNN attempts to perceive.}
On top of structures and features, CustomGNN, more importantly, can effectively perceive more fine-grained information, i.e., paths, and CustomGNN tends to extract high-level semantics of paths that are magnitude for different tasks. On the other hand, recently, lots of work focused on resolving the issue of over-smoothing for deep GNNs~\cite{chen2020simple,HE2023222,LI202250,DBLP:conf/icml/WuSZFYW19,zhu2021simple}.
CustomGNN provides a novel view to learn graph-structured data, which focuses on learning the high-level semantics of paths for specific downstream applications. \revision{This approach naturally addresses the aforementioned issues associated with traditional GNNs.}
\subsection{Graph Representation Learning}
Given a graph, node representation aims to learn the low-dimensional vector representation of each node and its graph-structured information. There have emerged many methods for node embedding in recent years. These methods can mainly be divided into two categories according to the input graph. \revision{The first category is sampling-based methods which use some sampling strategies, e.g., random walk~\cite{2016node2vec,2016Revisiting}, to sample subgraphs as the input data~\cite{chen2022variational,hamilton2017inductive}.} The advantage of these sampling-based methods is that they learn the local information of a graph, which can improve the model's generalization and reduce space consumption. The second category is inputting the whole graph~\cite{GCN,GAT}, and conducting specific GNNs on it to learn the node attributes and graph structure jointly. In addition, some advanced methods have emerged recently, for example, \citet{xu2021graphsad} disentangles the single representation of node attributes and graph structure into two different representations respectively. In this paper, we leverage the advantages of the sampling-based method to learn customized high-level semantic information, as well as the benefits of traditional GNNs to learn the generic graph information.
\subsection{Semi-Supervised Learning on Graph} 
One research direction for semi-supervised learning on graphs is to assign pseudo-labels to unlabeled data.
For instance, in order to utilize the unlabeled data, \citet{2013pesudo} first used a neural network to infer pseudo-labels of unlabeled data. \citet{yang2020rethinking} and \citet{202ASGN} performed the label propagation and pseudo-labels in the field of GNN. \citet{2015tirloss} used triplet loss to enhance the uniqueness of different faces for face recognition. Similarly, for semi-supervised learning, CustomGNN combined the pseudo-labels with triplet loss to enhance the semantics of each path.
A second research direction for semi-supervised learning on graphs is to design powerful regularization methods for regularizing GNNs~\cite{NEURIPS2020_fb4c835f,JUAN2023118935,10138449,rong2020dropedge}. For example, 
\revision{UGNN~\cite{10138449} utilized the unlabeled data by constructing them as subgraphs;}
GraphMix~\cite{2019GraphMix} introduced the mixup strategy~\cite{zhang2018mixup} in GNNs by utilizing linear interpolation between two nodes for data augmentation; GRAND~\cite{NEURIPS2020_fb4c835f} used the DropNode method (which is similar to DropEdge~\cite{rong2020dropedge}) to perturb the graph structure for data augmentation. Another branch of methods focuses on contrastive learning~\cite{MVGRL,InfoGraph,DGI,SMGCL}, proposing contrastive loss for optimizing among multiple views. For example, \citet{InfoGraph}, \citet{DGI} and \citet{MVGRL} tried to fuse the global graph information and local graph information by proposing different contrastive losses; \citet{SMGCL} injected label information in the contrastive learning process. These contrastive learning methods generally encouraged the embeddings to obtain more neglected information.

\revision{In contrast, CustomGNN approach integrates the multi-view concept from contrastive learning and incorporates the regularization idea, with a specific focus on semantics, aiming to enhance consensus among semantic representations.} Specifically, we perform random walks to generate various subgraphs in each iteration. Each of these subgraphs could represent a distinct view of the whole graph and each of the corresponding embeddings contains a distinct semantic. After multiple iterations, we get multi-perspective views and semantics. Finally, we compute the consistency loss among them to augment the consensus of multi-perspective semantics.

\section{The Proposed Method}
\label{Proposal}
\begin{figure*}[tb]
\centering
\includegraphics[width=1.0\textwidth]{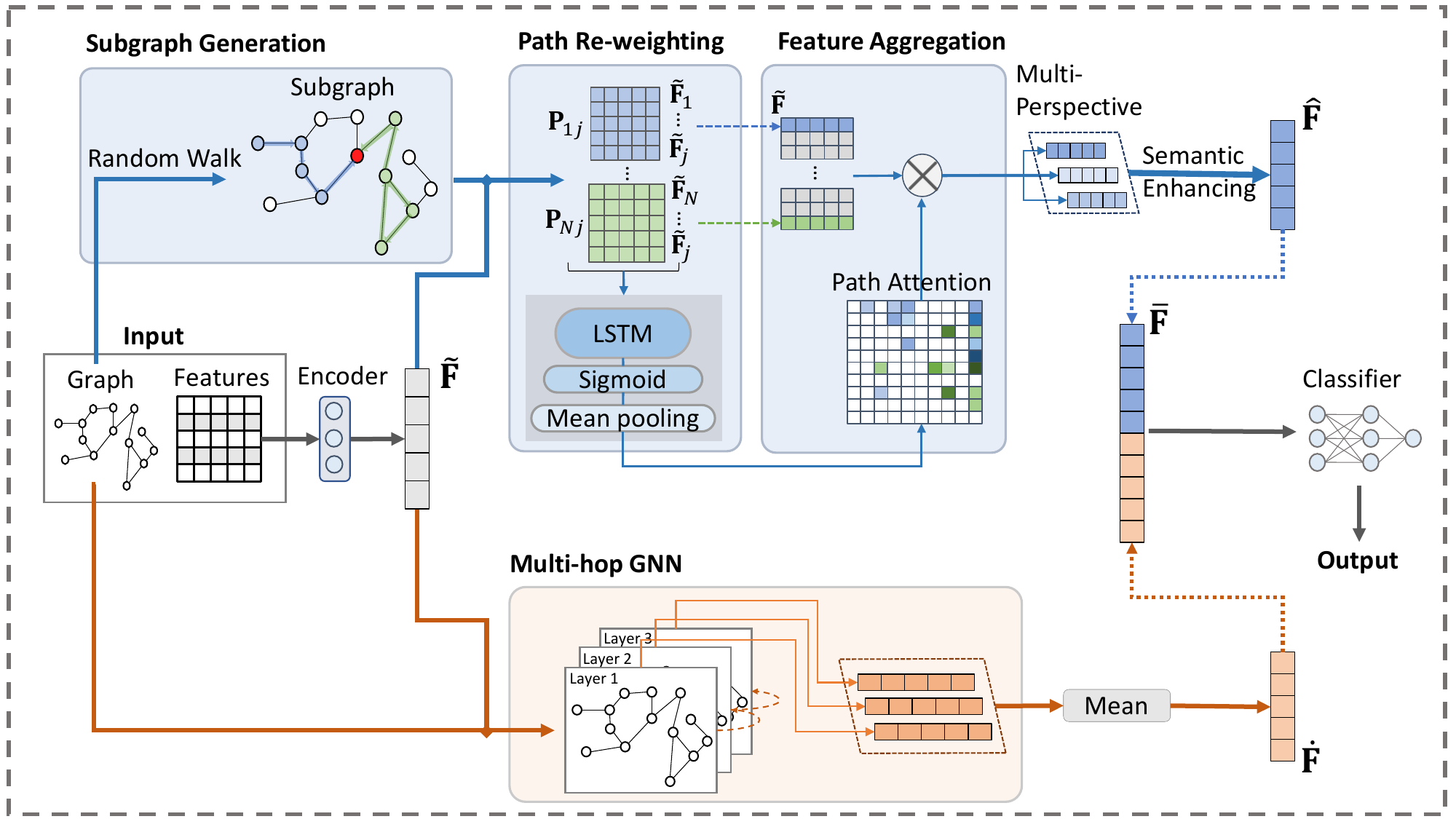}
\caption{The overall architectural diagram of CustomGNN. Here, $\otimes$ denotes matrix multiplication operation. There are two paths in CustomGNN. The first path highlighted in blue uses random walk to generate sub-paths (a subgraph), then, the path re-weighting module is utilized in these sub-paths for generating customized feature $\hat{\mathbf{F}}$. The orange line indicates that we utilize a simple multi-hop GNN to generate general feature $\dot{\mathbf{F}}$. Finally, we feed the aggregated feature $\{\overline{\mathbf{F}} \mid \overline{\mathbf{F}} = \hat{\mathbf{F}} \oplus \dot{\mathbf{F}}\}$ into classifier to obtain the predictions.}
\label{Architecture}
\end{figure*}

The overall architecture of CustomGNN is illustrated in Figure~\ref{Architecture}. Given a graph $G$ with encoded features, we develop a path re-weighting module (blue stream) to capture the customized high-level semantic information, and a multi-hop GNN module (orange stream) to capture the generic graph information. The most salient part proposed in CustomGNN is to re-weight each sub-path by an LSTM-based neural network model, which customizes the paths' importance for a specific downstream task.
Table~\ref{tab:notation} provides the definitions of the notations we used in this paper.

\begin{table}[tb]
    \centering
    \caption{The definitions of notations.}
    \begin{tabular}{ll}
    \toprule
        Notation & Definition
        \cr
    \midrule
          $\mathbf{A}$ & The adjacent matrix of the given graph $G$.\\
          $\mathbf{F}$ & The attributed features of the given graph $G$. \\ 
          $\widetilde{\mathbf{F}}$ & The output features from feature encoder $f_{enc}(\cdot)$ (Section~\ref{secFeatureEncoder}). \\
          $\hat{\mathbf{F}}$ & The task-oriented features from semantic aggregation (Section~\ref{sec:semanticAggregation}). \\
          $\dot{\mathbf{F}}$ & The generic features from multi-hop GNN module. \\
          $\overline{\mathbf{F}}^{(s)}$ &$ \overline{\mathbf{F}}^{(s)}=\dot{\mathbf{F}} \oplus \hat{\mathbf{F}}^{(s)}$ is the $s^{th}$ embeddings for multi-perspective regularization (Eq.~\eqref{eqLossCons}).\\
          $\overline{\mathbf{F}}$ & $\mathbf{F} = \hat{\mathbf{F}} \oplus \dot{\mathbf{F}}$ is the final concatenated embeddings. \\
          $\mathbf{P}^{(k)}_{ij}$ & $k^{th}$ path between node $i$ and node $j$. Each row in $\mathbf{P}$ is drawn from $\widetilde{\mathbf{F}}$.\\
          $\mathbf{W}$ & The path attention matrix computed by LSTM (Section~\ref{Path Re-weighting}).\\
          $\mathbf{Z}$ & The final predictions from classifier ($f_{mlp}(\cdot)$).\\
          $\boldsymbol{\theta}_1$ & The trainable parameters in encoder $f_{enc}(\cdot)$.\\
          $\boldsymbol{\theta}_2$ & The trainable parameters in LSTM ($f_{lstm}(\cdot)$).\\
          $\boldsymbol{\theta}_3$ & The trainable parameters in classifier ($f_{mlp}(\cdot)$).\\
          $\mathcal{L}_{sup}$ & The supervised loss computed from Eq.~\eqref{eqSupLoss}.\\
          $\mathcal{L}_{unsup}$ & The unsupervised loss consisted of $\mathcal{L}_{con}$ and $\mathcal{L}_{tri}$ (Eq.~\eqref{eqUnsupLoss}).\\
          $\mathcal{L}_{con}$ & The consistency loss for multi-perspective regularization computed from Eq.~\eqref{eqLossCons}.\\
          $\mathcal{L}_{tri}$ & The triplet loss for semantic enhancing computed from Eq.~\eqref{eqTriLoss}.\\
    \bottomrule
    \end{tabular}
    \label{tab:notation}
\end{table}

\subsection{Feature Encoder}
\label{secFeatureEncoder}
Let $n$ and $d$ represent the number of nodes and feature dimension respectively. In the beginning, a feature encoder $f_{enc}$ is employed to compress the sparse features $\mathbf{F} \in \mathbb{R}^{n \times d}$ into low dimensional dense features $\{\widetilde{\mathbf{F}}=f_{enc}(\mathbf{F}; \boldsymbol{\theta}_1) \mid \widetilde{\mathbf{F}} \in \mathbb{R}^{n \times D}\}$, where $D$ denotes the dimension of encoded features. The encoder $f_{enc}(\cdot)$ is implemented as: $f_{enc}(\mathbf{F}; \boldsymbol{\theta}_1)=\mathbf{F}\boldsymbol{\theta}_1$, where $\boldsymbol{\theta}_1 \in \mathbb{R}^{d \times D}$ are the trainable parameters in $f_{enc}(\cdot)$.
\subsection{LSTM-Based Path Re-Weighting}
\label{secCutomizedGAT}
\subsubsection{Subgraph Generation}
\label{paraPathSampling}
A random walk on the graph could generate multiple paths and extract corresponding node pairs. For example, setting the window size to three and path length to four, we can get a path $(x_1, x_2, x_3, x_4)$ and a set of sub-paths, i.e., $(x_1, x_2)$, $(x_2, x_3)$, $(x_3,x_4)$, $(x_1,x_2,x_3)$, $(x_2,x_3,x_4)$. The 
starting and ending node in each sub-path construct a node pair.
As illustrated in Figure~\ref{Architecture}, we perform multiple random walks from a single point, which will generate multiple paths with the same starting node. These paths with the same starting node can be regarded as a subgraph. Therefore, a subgraph can be generated by multiple random walks.
\subsubsection{Path Re-Weighting}
\label{Path Re-weighting}
In traditional GNNs such as GCN~\cite{GCN}, GCN\uppercase\expandafter{\romannumeral2}~\cite{chen2020simple}, SGC~\cite{DBLP:conf/icml/WuSZFYW19} and S\(^2\)GC~\cite{zhu2021simple}, paths were equally treated in their loss functions. Some other methods, such as GAT~\cite{GAT}, have computed the attention between a node pair. However, this attention is computed mainly according to the similarity of the node features, and the \textit{specific high-level semantic information} (such as the importance of paths between this pair of nodes) has been lost. \revision{Therefore, attention learned in this manner may not always be optimal for a given downstream task.} To better understand the importance of path re-weighting, we take the Wikipedia knowledge graph as an example. Each path in the graph reflects a piece of knowledge in a particular domain. If we want to enhance the performance of medical Q\&A by exploiting Wikipedia knowledge graph, some general knowledge that is not related to the medical domain should be de-emphasized. Thus, we re-weight the paths to indicate their importance scores so that the crucial knowledge for a specific task can be selected.

In the path re-weighting step, we generate path weight $w_{ij}$ for a node pair $(x_i, x_j)$. On account of homophily assumption~\cite{2001Homohoily}, nodes with the same label are more likely to be appeared in a same subgraph, so a subgraph must imply some semantic information about the labels. Therefore, the challenge is how to extract the implicit high-level semantics from subgraphs (paths).
Naturally, we can use a sequence model (e.g., LSTM~\cite{LSTM}) to extract the high-level semantic information from paths.

Formally, if we have a node pair $(x_i, x_j)$ and paths from $x_i$ to $x_j$, i.e., $\mathbf{P}_{ij}^{(0)}$, ... , $\mathbf{P}_{ij}^{(k)}$, ... , $\mathbf{P}_{ij}^{(N-1)}$, the path weight $w_{ij}$ can be computed by the followed formulation:
\begin{equation}
\label{eqPathReweighting}
    w_{ij} = \frac{1}{N} \sum_{k=0}^{N-1}  S(f_{lstm}(\mathbf{P}_{ij}^{(k)}; \boldsymbol{\theta}_2)),
\end{equation}
where $f_{lstm}$ and $\theta_2$ denote the LSTM model and the trainable parameters in LSTM respectively. $S(\cdot)$ represents Sigmoid activation function. The path matrix $\mathbf{P}_{ij} \in \mathbb{R}^{L \times D \times N}$ is the feature representation of sampled path sequences, where $L$, $D$ and $N$ represent the path length, dimension of $\widetilde{\mathbf{F}}$, and the number of paths between node pair $(x_i, x_j)$.

\subsection{Semantic Aggregation}
\label{sec:semanticAggregation}
\subsubsection{Re-Weighted Attention}
\label{paraGrapAttention}
We aggregate the semantic information from re-weighted paths in an attention matrix. 
Concretely, the path weights $\{w_{ij}|0 \leq i,j < n\}$ are used to construct path attention matrix $\mathbf{W} \in \mathbb{R}^{n \times n }$ whose role is similar to the attention matrix in GAT or the Laplacian matrix in GCN. Each weight value $w_{ij}$ is an entry of $\mathbf{W}$.

\subsubsection{Feature Aggregation}
\label{secGraphConvolution}
In this step, we conduct a convolution operation on $\mathbf{W}$ to embed features $\widetilde{\mathbf{F}}$. This operation could aggregate the high-level semantic information of a subgraph to center node ($j^{th}$ node) of this subgraph. The aggregating operation can be expressed as follow:
\begin{equation} 
\label{eqHatF}
  \hat{\mathbf{F}}_j = \sum_{i=0}^{n-1} w_{ij} \cdot \widetilde{\mathbf{F}}_{ij},
\end{equation}
where $w_{ij}$ is the attention (weight) value of $i^{th}$ node to $j^{th}$ node, $n$ represents the number of nodes, and $\hat{\mathbf{F}}_j$ denotes the  node embedding of $j^{th}$ node.

\subsection{Infusion with Multi-Hop GNN}
\label{secMultiGNN}
\subsubsection{Multi-Hop GNN}
To prevent some graph-structured information from being lost, the multi-hop GNN module is used to learn the generic graph information. In this module, we compute the means of all graph convolutional layers' output $\dot{\mathbf{F}}$. Formally, $\dot{\mathbf{F}}$ is computed as follow:
\begin{equation}
\label{eqMultiGNN}
    \dot{\mathbf{F}} = \frac{1}{H+1} \sum^{H}_{h=0} \mathbf{A}^h \widetilde{\mathbf{F}}, 
\end{equation}
where $H$ is a hyperparameter that denotes the number of graph convolutional layers.
\subsubsection{Semantic Infusion}
\label{paraSemanticInfuse}
Finally, we infuse the \textit{high-level semantic information} into generic graph information by concatenating $\hat{\mathbf{F}}$ and $\dot{\mathbf{F}}$, and feed them into an multi-layer perceptron (MLP) followed by a softmax function $\sigma(\cdot)$ to get the prediction $\mathbf{Z}$. The equation can be written as $\mathbf{Z} = \sigma (f_{mlp}(\dot{\mathbf{F}} \oplus \hat{\mathbf{F}} ; \boldsymbol{\theta}_3))$, where $\boldsymbol{\theta}_3$ is the trainable parameters in $f_{mlp}(\cdot)$, and $\oplus$ represents the concatenating operation.
\subsection{Loss Function and Joint Training}
\revision{In semi-supervised learning problem, the majority of training samples are unlabeled, so how to leverage these unlabeled data is important. Firstly, in the subgraph generation process (Section~\ref{paraPathSampling}), we randomly sample some subgraphs each iteration, these subgraphs can be seen as a new perspective of our data. Then, we propose multi-perspective regularization for regularizing these perspectives. Secondly, in order to enhance the distinctiveness among the generated path semantics, we first employ path re-weighting module to extract the semantic of each paths, and then, we minimize the distance of same semantics and maximize the distance of different one via triplet loss function. Therefore, the unsupervised loss consists of the two parts. Algorithm~\ref{agI} illustrates the whole training and predicting process. Next, we will introduce the training process in detail.}
\subsubsection{Overall Loss Function}
As formulated in Eq.~\eqref{overallLoss}, our loss function contains supervised and unsupervised loss, and we use $\lambda$ to control the trade-off between them:
\begin{equation}
\label{overallLoss}
    \mathcal{L} = \mathcal{L}_{sup} + \lambda \mathcal{L}_{unsup}.
\end{equation}

The unsupervised loss is a combination of the triplet loss (Eq.~\eqref{eqTriLoss}) and consistency loss (Eq.~\eqref{eqLossCons}) showed in Eq.~\eqref{eqUnsupLoss}, and we use two hyper-parameters $\lambda_1$ and $\lambda_2$ to control the balance of the two parts: 
\begin{equation} 
\label{eqUnsupLoss}
    \mathcal{L}_{unsup} = \lambda_1 \mathcal{L}_{con} + \lambda_2 \mathcal{L}_{tri}.
\end{equation}

With labeled data $\mathbf{Y}^{L}$ and the predictions $\mathbf{Z}^{(s),L}$ from MLP, we compute the average cross entropy loss of $\{\mathbf{Z}^{(s),L}|1 \leq s \leq S\}$:
\begin{equation}
\label{eqSupLoss}
    \mathcal{L}_{sup}=
    -\frac{1}{S}\sum^{S}_{s=1} [\mathbf{Y}^{L}\log{\mathbf{Z}^{(s),L}}+(1-\mathbf{Y}^L)\log{(1-\mathbf{Z}^{(s),L})}].
\end{equation}

\begin{algorithm}[!h]

\caption{CustomGNN}
\label{agI}
\textbf{Input}: adjacency matrix $\mathbf{A}$, feature matrix $\mathbf{F}$, training labels $\mathbf{Y}^L$;\\
\textbf{Hyperparameters}: loss trade-off parameters $\lambda$, $\lambda_1$ and $\lambda_2$, learning rate $\eta$, GNN propagation step $H$, the number of perspectives for regularization $S$;\\
\textbf{Parameters}: encoder $f_{enc}(\mathbf{F}; \boldsymbol{\theta}_1)$, sequence model (LSTM) $f_{lstm}(\mathbf{P}; \boldsymbol{\theta}_2)$, classifier $f_{mlp}(\overline{\mathbf{F}};\boldsymbol{\theta}_3)$.\\
\textbf{Output}: predictions $\mathbf{Z}$.
\begin{algorithmic}[1] 
\WHILE{not converged}
\STATE initialize $\mathbf{W}$ with $0$;\\
\FOR{s=0 to s \textless S}
\STATE Encode input features as Section~\ref{secFeatureEncoder};\\
\STATE Generate subgraphs for perspective $s$ by random walk;\\
\STATE Construct path attention matrix from sampled paths via Eq.~\eqref{eqPathReweighting};\\
\STATE Aggregate subgraph information via Eq.~\eqref{eqHatF};\\
\STATE Semantic enhancing via Eq.~\eqref{eqTriLoss};\\
\STATE Aggregate generic graph information via Eq.~\eqref{eqMultiGNN};\\
\STATE Semantic infusion and prediction as described in Section~\ref{paraSemanticInfuse}, i.e., $\mathbf{Z}^{(s)} = \sigma (f_{mlp}(\overline{\mathbf{F}}^{(s)}))$;
\ENDFOR
\STATE Compute the unsupervised loss $\mathcal{L}_{unsup}$ via Eq.~\eqref{eqUnsupLoss};\\
\STATE Compute the supervised loss $\mathcal{L}_{sup}$ via Eq.~\eqref{eqSupLoss};\\
\STATE Update the parameters $\boldsymbol{\theta}_1$, $\boldsymbol\theta_2$ and $\boldsymbol\theta_3$ by gradients descending: $\Theta = \Theta - \eta\nabla_{\Theta}(\mathcal{L}_{sup} + \lambda\mathcal{L}_{unsup})$;\\
\ENDWHILE
\STATE Output prediction $\mathbf{Z}$ as described in Section~\ref{paraPrediction}, i.e., $\mathbf{Z} = \sigma(f_{mlp}(\overline{\mathbf{F}}))$.
\end{algorithmic}
\end{algorithm}
\subsubsection{Multi-Perspective Regularization}
To improve robustness and prevent over-fitting, we propose multi-perspective regularization by improving the consistency regularization~\cite{berthelot2019mixmatch}, which encourages that different perspectives could get a consistent result. \revision{The main advantage is that the multiple perspectives are sampled from the input graph by random walk, which could capture and highlight different information so as to obtain a better regularization result.}

In doing so, we generate $S$ perspectives each epoch, each perspective is constructed by a separate path generation and re-weighting module (Section~\ref{secCutomizedGAT}), and produce $S$ different feature matrices $\{\hat{\mathbf{F}}^{(s)}|1 \leq s \leq S \}$. We concatenate $\hat{\mathbf{F}}^{(s)}$ with multi-hop GNN feature $\dot{\mathbf{F}}$ to get enhanced feature $\{\overline{\mathbf{F}}^{(s)} |\overline{\mathbf{F}}^{(s)}=\dot{\mathbf{F}} \oplus \hat{\mathbf{F}}^{(s)}\}$, the enhanced features $\overline{\mathbf{F}}^{(s)}$ are then fed into the MLP module to get the predictions $\{\mathbf{Z}^{(s)}|\mathbf{Z}^{(s)}=f_{mlp}(\overline{\mathbf{F}}^{(s)})\}$. The $S$ predictions $\{\mathbf{Z}^{(s)}|1 \leq s \leq S \}$ produced from $S$ perspectives are used for consistency regularization.

The purpose of this regularization is to minimize the $L_2$ distance among the predictions $\mathbf{Z}^{(s)}$. For example, setting $S=2$, we aim at minimizing ${\left \| \mathbf{Z}^{(1)} - \mathbf{Z}^{(2)} \right\|}^{2}_2$. To achieve this goal, we first need to obtain the average prediction for $i^{th}$ node, i.e., $\overline{\mathbf{Z}}_i = 1/S \sum_{s=1}^{S} \mathbf{Z}^{(s)}_i$. Then, we sharpen~\cite{berthelot2019mixmatch} the $\overline{\mathbf{Z}}_{i}$ with temperature $T$:
\begin{equation}
    \overline{\mathbf{Z}}_{ic}' = \frac{\overline{\mathbf{Z}}_{ic}^{1/T}} { \sum_{c=0}^{C-1}{\overline{\mathbf{Z}}_{ic}^{1/T}}} ,
\end{equation}
 where $\overline{\mathbf{Z}}_{ic}'$ is the sharpened average prediction of $i^{th}$ node in $c^{th}$ class. Finally, we compute the $L_2$ norm distance between the sharpened average prediction and $S$ individual predictions. The consistency loss can be written as:
\begin{equation} 
\label{eqLossCons}
    \mathcal{L}_{con} = \frac{1}{S} \sum\limits_{s=1}^{S} \sum\limits_{i=0}^{n-1}  {\left \| \mathbf{Z}_{i}^{(s)} - \overline{\mathbf{Z}}_{i}' \right\|}^{2}_{2}.
\end{equation}

\subsubsection{Semantic Enhancing with Triplet Loss}
Intuitively, the high-level semantics is implied in the labels for node classification task. Therefore, the essence of \textit{semantic enhancing} is to make the high-level semantic of paths in same label become close and different labels become distant.
\begin{equation}
\label{eqI}
\mathbf{I}_{ij}=
    \left\{
    \begin{array}{lr}
         1, &  \mathbf{Y}_{i} = \mathbf{Y}_{j},\\
         0, &  \mathbf{Y}_{i} \neq \mathbf{Y}_{j}.
    \end{array}
    \right.
\end{equation}
\revision{The matrix $\mathbf{I} \in \{0,1\}^{n \times n}$ is introduced to indicate whether the labels are the same or not.} In Eq.~\eqref{eqI}, $\mathbf{Y} \in \{0,1\}^{n \times C}$ denotes the one-hot label vector, $C$ represents the number of classes. Notably, the labeled $i^{th}$ node $\mathbf{Y}_i \in \mathbf{Y}^L$ is obtained from training set directly, and the unlabeled $j^{th}$ node $\mathbf{Y}_j \in \mathbf{Y}^U$ is ``guessed'' by our model to get a pseudo label. The pseudo labels can be generated by $\{\mathbf{Y}_j = argmax(\mathbf{Z}_j)|\mathbf{Y}_j \in \mathbf{Y}^U \}$, where $\mathbf{Z}$ denotes the probabilities predicted by CustomGNN. Then, the positive node pairs $(\hat{\mathbf{F}}_{i}^{pos}, \hat{\mathbf{F}}_{j}^{pos})$ and negative node pairs $(\hat{\mathbf{F}}_{i}^{neg}, \hat{\mathbf{F}}_{j}^{neg})$ can be sampled out. We use triplet loss~\cite{2015tirloss} to minimize the $L_2$ norm distance between positive node pairs and maximize the distance between negative node pairs. The loss function is as follows:
\begin{equation}
\label{eqTriLoss}
    \mathcal{L}_{tri} =  {\left \| \hat{\mathbf{F}}_{i}^{pos} - \hat{\mathbf{F}}_{j}^{pos} \right\|}_2 + RELU(m - {\left \| \hat{\mathbf{F}}_{i}^{neg} - \hat{\mathbf{F}}_{j}^{neg} \right\|}_2),
\end{equation}
where $m$ is a hyperparameter that represents the margin between negative node pairs. 

\subsubsection{Inference} 
\label{paraPrediction}
To avoid the path attention matrix $\mathbf{W}$ being too sparse we compute this matrix $S$ times ($S=4$ in our setting) by $S$ perspectives and add them together to get a new path attention matrix $\{\mathbf{W}| \mathbf{W} = \sum_{s=1}^S \mathbf{W}^{(s)}\}$. The final path attention matrix is used to compute $\hat{\mathbf{F}}$ as Eq.~\eqref{eqHatF}. Meanwhile, $\dot{\mathbf{F}}$ is computed by multi-hop GNN module. At last, the concatenated embedding $\overline{\mathbf{F}}$ is fed into classifier, i.e., $f_{mlp}(\cdot)$ to get the final prediction $\mathbf{Z}$.

\subsection{Complexity Analysis}
CustomGNN consists of four main components: subgraph generation, path re-weighting, feature aggregation, and multi-hop GNN. To analyze the time complexity of the model, we need to consider the time complexity of each component and sum them up.
\begin{itemize}
    \item Subgraph generation: this component uses random walk to generate subgraphs from the original graph. Assume $n$ is the number of nodes, $k$ is the iterations of random walk, so the time complexity of this component is $O(kn)$.
    \item Path re-weighting: this component uses a LSTM-based path weighting strategy to calculate multiple path attention maps from the subgraphs. Assume $l$ is the average length of each path, and $h$ is the hidden size of LSTM, so for average $p$ paths in each subgraph, the time complexity of this component is $O(plh^2)$. Notably, LSTM can be replaced by other more efficient sequence models, e.g., GRU (Gate Recurrent Unit) and self-attention neural networks.
    \item Feature aggregation: this component uses a weighted sum operation to aggregate features through the path attention maps. The time complexity of this component is $O(nd)$, where $d$ is the dimension of node features.
    \item Multi-hop GNN: this component uses a standard multi-hop GNN to infuse the general structural information. The time complexity of this component is $O(onmd)$, where $o$ is the number of hops or layers, and $m$ is the average degree of nodes in the original graph.
\end{itemize}
Overall, the training time complexity of CustomGNN is $O(kn + plh^2 + nd + onmd)$. In the inference phase, the path attention map is directly loaded from saved parameters. Therefore, inference time complexity is $O(nd + onmd)$, which is approximately the same as other popular GNNs.

\section{Experiments}
\label{experiments}
\subsection{Datasets}
Following the community convention, we use three benchmark graphs, i.e., Cora, Citeseer and PubMed with their standard public splits in Planetoid~\cite{2016Revisiting}, which contains 20 nodes per class for training and 1000 nodes for testing. We also did experiments on four publicly available large datasets, i.e., Cora Full, Coauthor CS, Amazon Photo, and Amazon Computers with their experimental settings in \citet{2018Pitfalls}. 

These datasets can be downloaded from PyTorch-Geometric library\footnote{\url{https://pytorch-geometric.readthedocs.io}}. The datasets details are shown in Table~\ref{tab:datasetDetail}.
Label rate is the fraction of nodes in training set (the number of training nodes is 20 per class) that can be computed as (\#class $\cdot$ 20)/\#node.
\begin{table*}[tb]
\begin{center}
  \caption{Details of datasets after preprocessing.}
  \newcommand{\minitab}[2][l]{\begin{tabular}{#1}#2\end{tabular}}
  \begin{tabular}{lrrrrr}
    \toprule
    Dataset & Nodes  & Edges  & Classes  & Features & Label Rate   
    \cr
    \midrule
    Cora & 2,708 & 5,429 & 7 & 1,433 & 0.0516  \tabularnewline
    Citeseer & 3,327 & 4,732 & 6 & 3,703 & 0.0360 \tabularnewline
    PubMed & 19,717 & 44,338 & 3 & 500 &  0.0030 \tabularnewline
    Cora-Full & 19,749 & 63,262 & 68 & 8,710 & 0.0689 \tabularnewline
    Coauthor CS & 18,333 & 81,894 &15 &6,805  & 0.0164 \tabularnewline
    Amazon Computer & 13,752 & 245,861 &10  & 767 & 0.0145 \tabularnewline
    Amazon Photo  &7,650  & 119,081 & 8  & 745 & 0.0209  \tabularnewline
    \bottomrule
  \end{tabular}
  \label{tab:datasetDetail}
\end{center}
\end{table*}

\subsection{Baselines} 
For the three standard split datasets, we choose 20 GNN SOTAs for comparison, including GCN~\cite{GCN} and its evolutions \cite{chen2020simple,DBLP:conf/icml/WuSZFYW19,zhu2021simple}, deep GNNs like MixHop~\cite{zhang2018mixup} and APPNP~\cite{DBLP:conf/iclr/KlicperaBG19}, and five augmentation-based GNNs~\cite{NEURIPS2020_fb4c835f,rong2020dropedge,2019GraphMix,yang2020rethinking}.
For the other four larger datasets, we choose baseline methods that can be scaled to large graphs, including a two-layer MLP (input layer $\rightarrow$ hidden layer $\rightarrow$ output layer) with 128-hidden units, standard GCN and GAT, and two regularization based methods, i.e., GRAND~\cite{NEURIPS2020_fb4c835f} (using MLP as the backbone) and P-reg~\cite{yang2020rethinking} (using GCN as the backbone).

\subsection{Overall Results}
\begin{table}[]
   \caption{Results on 3 standard split datasets with over 100 runs. Some papers do not report their std values, which are marked by ``*''. \textbf{Bold} denotes the best performance, \underline{underline} denotes the second best.}
    \centering
   \begin{tabular}{rlll}
    \toprule
    Method & Cora & Citeseer & PubMed\cr  
    \midrule
    GCN (\citeyear{GCN}) & 81.5\(\pm\) * & 70.3\(\pm\) * & 79.0\(\pm\) * \tabularnewline
    GraphSAGE (\citeyear{hamilton2017inductive}) & 78.9\(\pm\)0.8 & 67.4\(\pm\)0.7 &  77.8\(\pm\)0.6\tabularnewline
    FastGCN (\citeyear{Chen2018fastGCN})& 81.4\(\pm\)0.5 & 68.8\(\pm\)0.9 &  77.6\(\pm\)0.5\tabularnewline
    GAT (\citeyear{GAT}) & 83.0\(\pm\)0.7 & 72.5\(\pm\)0.7 & 79.0\(\pm\)0.3\tabularnewline
    MixHop (\citeyear{zhang2018mixup}) & 81.9\(\pm\)0.4 & 71.4\(\pm\)0.8 & 80.8\(\pm\)0.6\tabularnewline
    APPNP (\citeyear{DBLP:conf/iclr/KlicperaBG19}) & 83.8\(\pm\)0.3 & 71.6\(\pm\)0.5 & 79.7\(\pm\)0.3\tabularnewline
    DropEdge (\citeyear{rong2020dropedge}) & 82.8\(\pm\) * & 72.3\(\pm\) *  & 79.6\(\pm\) * \tabularnewline
    SGC (\citeyear{DBLP:conf/icml/WuSZFYW19}) & 81.0\(\pm\)0.0 & 71.9\(\pm\)0.1 & 78.0\(\pm\)0.0\tabularnewline
    GMNN (\citeyear{DBLP:conf/icml/QuBT19}) & 83.7\(\pm\) * & 72.9\(\pm\) * & 81.8\(\pm\) *\tabularnewline
    GraphNAS (\citeyear{DBLP:journals/corr/abs-1904-09981}) & 84.2\(\pm\)1.0 & 73.1\(\pm\)0.9 & 79.6\(\pm\)0.4\tabularnewline
    GCN\uppercase\expandafter{\romannumeral2} (\citeyear{chen2020simple}) &\textbf{85.5\(\pm\)0.5} & 73.4\(\pm\)0.6 & 80.2\(\pm\)0.4\tabularnewline
    GRAND (\citeyear{NEURIPS2020_fb4c835f}) & \underline{85.4\(\pm\)0.4} & \underline{75.4\(\pm\)0.4} & \underline{82.7\(\pm\)0.6} \tabularnewline
    GraphMix (\citeyear{2019GraphMix}) & 83.9\(\pm\)0.6 & 74.5\(\pm\)0.6 & 81.0\(\pm\)0.6\tabularnewline
    P-reg (\citeyear{yang2020rethinking}) & 82.8\(\pm\)1.2 & 71.6\(\pm\)2.2 & 77.4\(\pm\)1.5\\
    superGAT (\citeyear{2021superGAT}) & 84.3\(\pm\)0.6 & 72.6\(\pm\)0.8 & 81.7\(\pm\)0.5    \tabularnewline
    S\(^2\)GC (\citeyear{zhu2021simple}) & 83.5\(\pm\)0.0 & 73.6\(\pm\)0.1 & 80.2\(\pm\)0.0\tabularnewline
    GraphSAD (\citeyear{xu2021graphsad})& 83.0\(\pm\)0.4 & 71.2\(\pm\)0.2 & 79.6\(\pm\)0.1\tabularnewline
    CGNN (\citeyear{LI202250}) & 82.5\(\pm\)0.6  & 72.1\(\pm\)0.7 & 78.9\(\pm\)0.5\tabularnewline
    \midrule
    \textbf{CustomGNN} & \textbf{85.5\(\pm\)0.2} & \textbf{76.0\(\pm\)0.4} & \textbf{83.3\(\pm\)0.2}
    \tabularnewline
    \bottomrule
   \end{tabular}
    \label{tab:simpleOverall}
\end{table}

\begin{table}[tb]
\caption{Results on 4 large datasets with over 100 runs on different random seed for train/test set split and different random seed for weight initialization. \textbf{Bold} denotes the best performance, \underline{underline} denotes the second best and ``--'' denotes the original authors does not run this experiment.}
\begin{center}
  \newcommand{\minitab}[2][l]{\begin{tabular}{#1}#2\end{tabular}}
  \begin{tabular}{rcccc}
    \toprule
    Method   & Cora-Full & Coauthor CS  &  Amazon Computer & Amazon Photo   
    \cr
    \midrule
    MLP & 18.3\(\pm\)4.9 & 88.3\(\pm\)0.7                     & 57.8\(\pm\)7.0  & 73.5\(\pm\)9.7 \tabularnewline
    GCN (\citeyear{GCN}) & 9.0\(\pm\)4.9  & 88.1\(\pm\)2.7                     & 42.3\(\pm\)16.0 & 61.4\(\pm\)12.3 \tabularnewline
    GAT (\citeyear{GAT}) & 13.7\(\pm\)5.2  & 90.3\(\pm\)1.4                   & 65.4\(\pm\)15.3 & 80.8\(\pm\)9.5 \tabularnewline
    GRAND (\citeyear{NEURIPS2020_fb4c835f}) & \underline{42.3\(\pm\)6.5}  & \underline{92.9\(\pm\)0.5}  & 80.1\(\pm\)7.2  & 72.9\(\pm\)2.1\tabularnewline
    P-reg (\citeyear{yang2020rethinking}) &-- & 92.6\(\pm\)0.3 & \underline{81.7\(\pm\)1.4} & \underline{91.2\(\pm\)0.8}  \tabularnewline
    \midrule
    \bf{CustomGNN}& \bf{44.2\(\pm\)1.2} & \bf{93.4\(\pm\)0.3}  & \bf{81.9\(\pm\)0.8} & \bf{92.0\(\pm\)3.4} \tabularnewline
    \bottomrule
  \end{tabular}
  \label{tab:largeDataset}
\end{center}
\end{table}

Table~\ref{tab:simpleOverall} compares the accuracy of CustomGNN with 21 SOTA methods. The results of these methods are taken from their original papers directly. It shows that CustomGNN achieves the SOTA accuracy on all three standard datasets, i.e., Cora, Citeseer, and PubMed. \revision{It is notable that CustomGNN shows impressive improvements on Citeseer, which may be because the citation graph on Citeseer implies more semantics, and CustomGNN has effectively extracted and utilized this semantic information. We will further analyze this finding in Section~\ref{Analysis}.}

Table~\ref{tab:largeDataset} compares the accuracy of CustomGNN with three traditional baselines and two regularization based GNNs. The experiments strictly follow the evaluation protocol of \citet{2018Pitfalls}. The results of the MLP, GCN, GAT, GRAND and CustomGNN are averaged over 100 runs on different random seed for train/test set split and different random seed for weight initialization. The results of P-reg are taken from its original paper. It shows that CustomGNN outperforms the other five methods on all four large datasets.

\subsection{Ablation Study}
Table~\ref{tab:ablation} illustrates the results of ablation study. This study evaluates the contributions of different components in CustomGNN.
\begin{table}[tb]
\centering
\caption{Ablation studies. The accuracy of CustomGNN without specific component. The number in parentheses is the decay of accuracy after removing this component.}
        \begin{tabular}{lccc}
            \toprule
            Component & Cora  & Citeseer & PubMed
            \cr
            \midrule
             w/o path re-weighting   & 79.2(-6.3) & 68.7(-7.3)  & 77.8(-5.5) \\
             w/o multi-hop GNN   & 83.4(-1.2) & 75.4(-0.6)  & 81.6(-1.7) \\
             w/o semantic enhancing & 84.6(-1.2) & 75.2(-0.8)  & 81.6(-1.7) \\
             w/o multi-perspective & 84.2(-1.3) & 75.3(-0.7)  & 79.2(-4.1) \cr 
            \midrule
             original & \textbf{85.5} & \textbf{76.0}  & \bf{83.3}
             \cr
            \bottomrule
        \end{tabular}
    
    \label{tab:ablation}
\end{table}
\revision{
\paragraph{Without Path Re-Weighting.}
In this part, we only use the multi-hop GNN module to generate node embeddings, i.e., $\overline{\mathbf{F}}=\dot{\mathbf{F}}$, the path re-weighting module containing unsupervised loss (Eq.~\eqref{eqUnsupLoss}) is removed. So the learned embedding only contains generic graph information. This result can be regarded as a baseline.
\paragraph{Without Multi-Hop GNN.}
This ablation experiment only uses the path re-weighting module to generate the final node embeddings, i.e., $\overline{\mathbf{F}}=\hat{\mathbf{F}}$. In addition, to get a dense path attention matrix ($\mathbf{W}$), we sample paths for all nodes, i.e., fixing sampling batch size to \#nodes/S. It is obvious that the classification results are better than w/o path re-weighting, indicating that the path re-wieghting module learns more effective information than generic graph information learned by multi-hop GNN.
\paragraph{Without Semantic Enhancing.}
The unsupervised loss is only computed by consistency regularization loss to show the efficacy of the proposed semantic enhancing method with triplet loss. The results indicate that the accuracy exhibits a slight decline when the semantic enhancing regularization term is omitted.
\paragraph{Without Multi-Perspective Regularization.}
The unsupervised loss is only computed by triplet loss to show the efficacy of the proposed multi-perspective regularization. To this end, we first set $S=1$ for using only one perspective. Second, we sample paths for all nodes to ensure that this perspective can get more complete information. Third, because there is no consistency regularization, we decrease the dropout rate to avoid overfitting. The results show that the exclusion of the multi-perspective regularization term evidently leads to a decline in the results, particularly when dealing with relatively larger datasets, i.e., PubMed.
}
\subsection{Analysis}
\label{Analysis}
\begin{figure}[tb]
\centering
    \subfigure[Distribution of re-weighted path attentions.]{
    \includegraphics[width=0.32\linewidth]{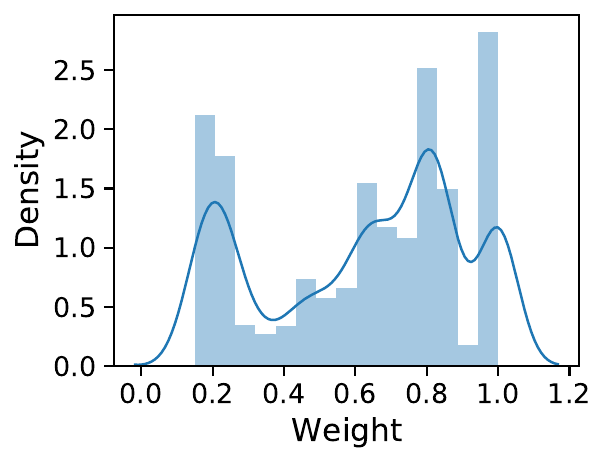}
    \label{fig:visWeights}
    }
    \subfigure[Visualization of relevances among categories.]{
    \includegraphics[width=0.33\linewidth]{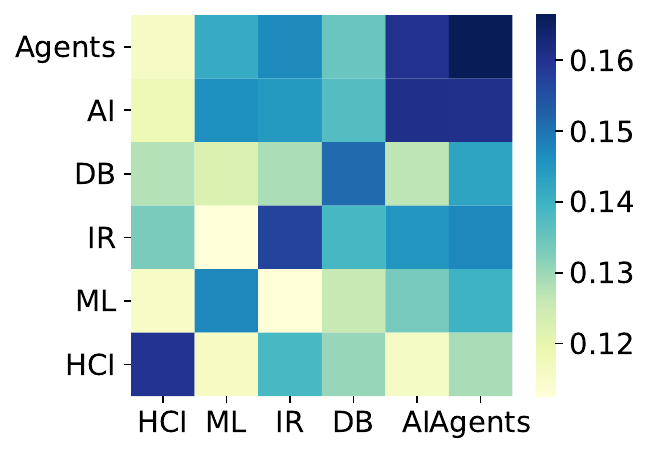}
    \label{fig:semanticExplanation}
    }
\caption{(a) The visualization of re-weighted path attentions on Citeseer. We filter out some noisy points whose weight is less than 0.15. (b) The visualization of relevance between different categories on Citeseer. The weight on $(i,j)$ means how much the $i^{th}$ category is influenced by $j^{th}$ category.}
\end{figure}

The re-weighted attentions learned by CustomGNN are customized for different downstream tasks. For example, Figure~\ref{fig:visWeights} shows the distribution of re-weighted attentions for Citeceer node classification task, which is quite diverse. To better understand these attentions and how CustomGNN benefits the results, we provide some semantic analyses. After that, we study three metrics of our performance, i.e., generalization, robustness, and over-smoothing. In our experiments, we compared CustomGNN with two traditional GNNs (GCN~\cite{GCN} and GAT~\cite{GAT}) and a GNN SOTA (GRAND~\cite{NEURIPS2020_fb4c835f}) which concentrates on resolving the three issues.
\subsubsection{Semantics Explanation}
\label{sec:SemanticExp}
\begin{table}[tb]
    \centering
    \caption{Summary of paper citation graph on Citeseer.}
    \begin{tabular}{cc}
        \toprule
         Category & Number
         \cr
         \midrule
         Agents & 239\\
         Artificial Intelligence (AI) & 537\\
         Data Base (DB) & 619\\
         Information Retrieval (IR) & 633 \\
         Machine Learning (ML) & 560\\
         Human–Computer Interaction (HCI) & 463\\
        \bottomrule
    \end{tabular}
    \label{tab:classInfo}
\end{table}
In order to mine the semantics of re-weighted paths, we use Citeseer as an example. Papers on Citeseer are classified into 6 categories as described in Table~\ref{tab:classInfo}. Based on this dataset, the task is to classify a given paper to its category. After training the model, we generate multiple paths for a pair of nodes, the average of these paths' weights is the average relevance weight between the corresponding categories of this pair of nodes.
We visualize the relevance weights among different categories in Figure~\ref{fig:semanticExplanation}.

Figure~\ref{fig:semanticExplanation} implies a lot of information. Consistent with our intuition, the average weights of self-loop (computed from paths where the starting and ending nodes are from the same category) is the largest, e.g., the weight of (\textit{IR,IR}) is larger than (\textit{IR, others}). Moreover, from Figure~\ref{fig:semanticExplanation}, we can see that \textit{AI} and \textit{Agents} are highly relevant with each other; \textit{AI} is much influenced by \textit{ML}, whereas \textit{ML} is less influenced by \textit{AI}; \textit{DB} influences others more than it is influenced by others, as the column of \textit{DB} has smaller average weights than its corresponding line in the visualized matrix. All these results coincide with human intuitions.

\subsubsection{Weight Correlation}
\label{sec:weightCorrelation}
\begin{figure}[tb]
\centering
        \subfigure[Weight-Length on Cora.]{
        \includegraphics[width=0.23\linewidth]{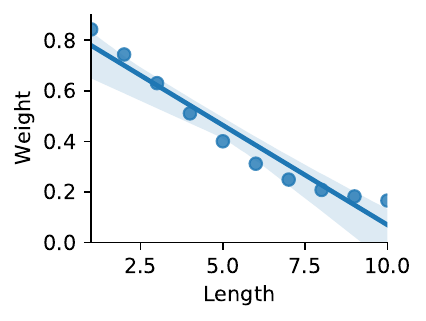}
        \label{weightLengthCora}
        }
        \subfigure[Weight-Diversity on Cora.]{
        \includegraphics[width=0.23\linewidth]{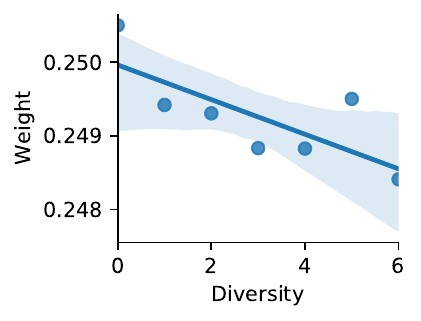}
        \label{weightDiversityCora}
        }
     \subfigure[Weight-Length on Citeseer.]{
    \includegraphics[width=0.23\linewidth]{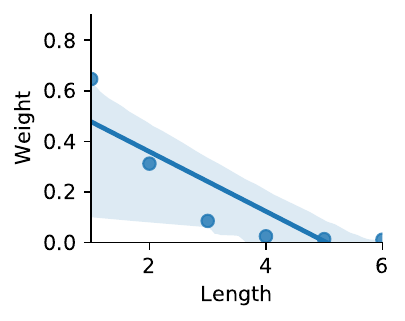}
    \label{weightLengthCiteseer}
    }
    \subfigure[Weight-Diversity on Citeseer.]{
    \includegraphics[width=0.23\linewidth]{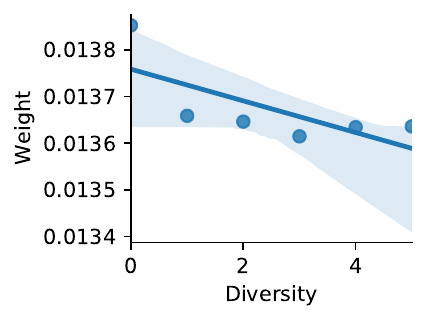}
    \label{weightDiversityCiteseer}
    }
\caption{(a), (b) and (c), (d) are the mean correlations between weight and two properties (length and diversity) of paths on Cora and Citeseer respectively. In (b) and (d), to remove the influence of length, the path length is fixed on 10. Y-axis denotes the average weight of the corresponding paths.}
\label{weightPath}
\end{figure}
To analyze how the properties of paths influence weight scores, we draw the correlations between weight scores and two properties of paths, i.e., length and diversity, on Figure~\ref{weightPath}. Length is defined as the number of nodes in a path; Diversity is defined as the number of different categories of labeled node in a path. For example, we have a path ($x_1, x_2, x_3$) and their corresponding categories (\textit{AI}, \textit{Agents}, \textit{AI}), the length and diversity of this path is three and two respectively. As demonstrated in Figures~\ref{weightLengthCora} and \ref{weightLengthCiteseer}, correlations between two nodes decay as the distance between them becomes larger, this corresponds with the homophily assumption~\cite{2001Homohoily} that the correlation is strong if two nodes are adjacent; In Figures~\ref{weightDiversityCora} and \ref{weightDiversityCiteseer}, the path weight increases with the reduction of node diversity. Intuitively, if a path contains fewer categories, the path contains more information and less noise about the corresponding class. This nature is helpful for the task of node classification.

\subsubsection{Generalization Analysis}
\begin{figure}[tb]
\centering
        \subfigure[W$\backslash$o re-weigting.]{
        \includegraphics[width=0.25\linewidth]{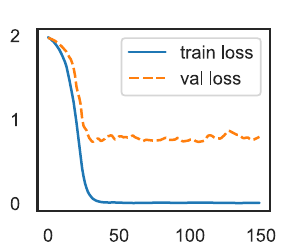}
        }
        \subfigure[CustomGNN.]{
        \includegraphics[width=0.25\linewidth]{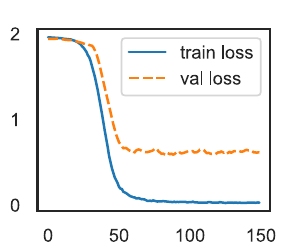}
        }
        \subfigure[Gap.]{
        \includegraphics[width=0.25\linewidth]{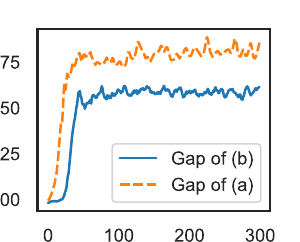}
        }
\caption{(a) is the loss of CustomGNN without path re-weighting, and (b) is the loss of CustomGNN. X-axis denotes training epochs, Y-axis denotes loss values. (c) illustrates the gap between train loss and validation loss of (a) and (b), Y-axis denotes the gap value.}
    \label{figLoss}
\end{figure}

We examine the generalization of CustomGNN and how the path re-weighting module contributes to the generalization. To measure this, we compute the gap of cross-entropy losses between train set and validation set. The smaller gap illustrates the better generalization ability of the model. Figure \ref{figLoss} reports the results, and proves that the techniques of path re-weighting module in CustomGNN can prevent overfitting.

\subsubsection{Robustness Analysis}
\begin{figure}[tb]
\centering
    \subfigure[Robustness analysis.]{
    \includegraphics[width=0.25\linewidth]{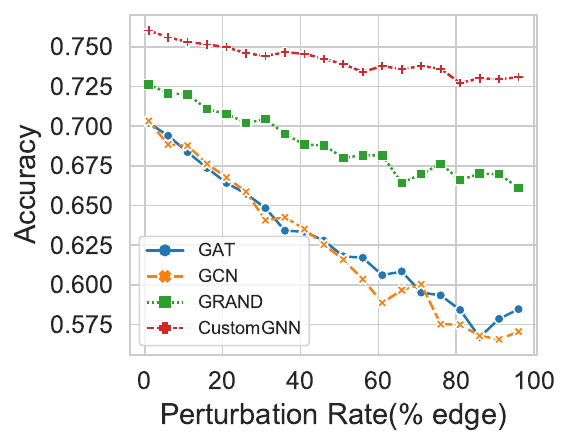}
    \label{figRob}
    }
    \subfigure[Over-smoothing analysis.]{
    \includegraphics[width=0.24\linewidth]{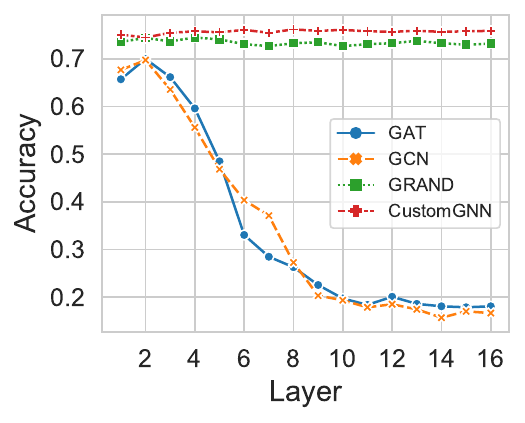}
    \label{figOverSmoothing}
    }
    \subfigure[Dimentional reduction analysis on Cora.]{
    \includegraphics[width=0.225\linewidth]{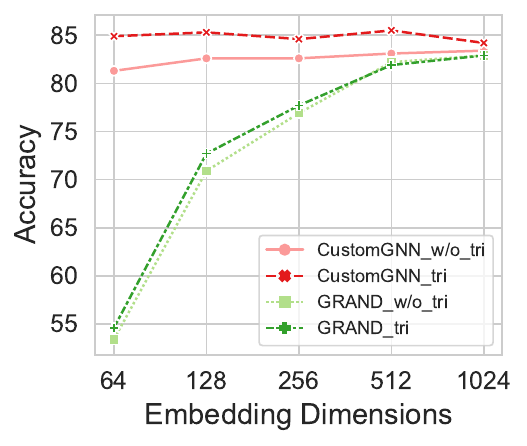}
    \label{figDimCora}
    }
    \subfigure[Dimentional reduction analysis on Citeseer.]{
        \includegraphics[width=0.225\linewidth]{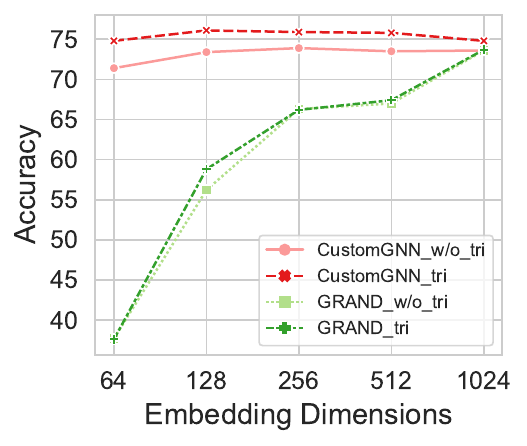}
    \label{figDimCiteseer}
    }
\caption{(a) On Robustness analysis, we gradually increase the perturbation rate (from 1\% to 100\%). (b) On over-smoothing analysis, for GAT, GCN, and GRAND, we increase the propagation layers; For CustomGNN, we increase the propagation layers of multi-hop GNN module, and increase the aggregation orders (window size) of the path re-weighting module meanwhile. (c) and (d) are dimensional reduction analysis, where we generally increase the embedding dimensions from 64 to 1024. }
\end{figure}
We made a noise identification ability and robust analysis of CustomGNN by randomly adding a certain proportion of fake edges. In Figure \ref{figRob}, we can observe that the classification accuracy of nodes in GCN and GAT declines rapidly with the increase of fake edges. Although both CustomGNN and GRAND can maintain a slower decline rate in classification accuracy as the number of fake edges increases, CustomGNN is always superior to GRAND in robustness, and the decline curve is smoother.

\subsubsection{Over-Smoothing Analysis}
Lots of GNN models suffer from the over-smoothing problem. As reported in previous works~\cite{zhu2021simple,2019Measuring}, with the increase of layer numbers, the identification of nodes in different classes becomes undistinguished, because the multiple propagation steps lead to over-mixing of information and noises~\cite{2019Measuring}, then, the node embeddings become similar. Figure~\ref{figOverSmoothing} proved the ability of our model to relieve over-smoothing.

\subsection{Dimensional Reduction Analysis}
In reality, many nodes attach some auxiliary information, for example, each node on Cora-Full attaches a feature with 8710-dimension. Therefore, the high dimensional features expect low dimensional representations. In this experiment, we conduct encoder $f_{enc}$ to reduce the feature dimension in the beginning, then, CustomGNN is compared with GRAND in different dimensions. As shown in Figures~\ref{figDimCora} and \ref{figDimCiteseer}, with the decrease of dimension, the accuracy of GRAND declining, by contrast, the accuracy of CustomGNN is stable. The result demonstrates that CustomGNN is not sensitive to different embedding dimensions. A possible reason is that CustomGNN can extract the task-relevant information more efficiently.
Alternatively, from Figures~\ref{figDimCora} and \ref{figDimCiteseer}, it is notable that the triplet loss which is used for semantic enhancing can benefit more to CustomGNN than GRAND. This might be that the CustomGNN captures more semantic information that has the chance to be enhanced by the proposed triplet loss.

\section{Conclusions}
\revision{In this work, we study graph neural networks (GNN) from a new view, i.e., customizing GNN for a specific downstream task, and present a new GNN framework, namely, Customized Graph Neural Network with Path Reweighting (CustomGNN). CustomGNN can resolve the inherent issues of traditional GNNs, i.e., over-smoothing, over-fitting, and non-robustness to data attack. In CustomGNN, we capture customized semantic information from weighted paths, which is then infused with the generic graph information extracted from multi-hop GNN module. Moreover, based on the well-designed framework, we generate multi-perspective subgraphs to regularize the extracted semantics, and enhance the semantics by proposing a triplet loss with pseudo-labels. Extensive experiments show that CustomGNN outperforms most SOTAs. In addition, we analyze the semantics learned by CustomGNN and demonstrate the superiority of CustomGNN in terms of resistance to over-smoothing and robustness to data attack.
In future work, we aim to apply the impressive semantic extraction ability of CustomGNN to more graph-based tasks.}

\printcredits

\section*{Declaration of Competing Interest}
The authors declare that they have no known competing financial interests or personal relationships that could have
appeared to influence the work reported in this paper.
\section*{Acknowledgements}
This work was supported in part by National Key Research and Development Program of China (No. 2020YFB2103402), Shenzhen Science and Technology Program (No. JCYJ20230807115959041), and the open project of Sichuan Provincial Key Laboratory of Philosophy and Social Science for Language Intelligence in Special Education (No. YYZN-2023-3).

\appendix

\section{Reproducibility}
\label{Reproducibility}
\subsection{Implementation Details}
We use PyTorch to implement CustomGNN and all of its components. The LSTM we used in path re-weighting module is implemented in the package of torch.nn.LSTM. For the results of large datasets we reported (in Table~\ref{tab:largeDataset}), the implementation of GRAND comes from its public source code\footnote{\url{https://github.com/Grand20/grand}}, the implementations of GCN and GAT layer come from the PyTorch-Geometric library, and the results of P-reg are taken from its original paper directly. We adopt Adam to optimize the parameters of all models in our paper, and we perform early stopping strategy to control the training epochs. We employ Dropout in the adjacency matrix, path-weight matrix, encoder, and each layer of prediction module (i.e., MLP) as a generally used trick for preventing overfitting. The experiments of Cora, Citeseer are conducted on a Tesla V100 with 80GB memory size, the experiments of PubMed, Cora Full, Amazon Computer, Amazon Photo and Cauthor CS are conducted on Tesla V100 with 32GB memory size. As for the software version, we use Python 3.8.5, PyTorch 1.7.1, NumPy 1.19.2, and CUDA 11.0.

\subsection{Hyperparameter Details}
We show the hyperparameters of CustomGNN for results in Table~\ref{tab:simpleOverall}. These hyperparameters can be divided into 4 groups, the first controls the training process which is shown in 1 to 12 rows of Table~\ref{Hyperparameters}, the second controls the customized attention module which is shown in 13 to 16 rows of Table~\ref{Hyperparameters}, the third controls the triplet loss which are shown in 17 to 20 rows of Table~\ref{Hyperparameters}, and the fourth control the the consistency loss which are shown in 21 to 22 rows of Table~\ref{Hyperparameters}.
\begin{table}[!h]
    \caption{Hyperparameters of CustomGNN for reproducing the results reported in Table~\ref{tab:simpleOverall}.}
    \label{Hyperparameters}
    \centering
    \begin{tabular}{rccc}
    \toprule
    Hyperparameters & Cora & Citeseer & PubMed \\
    \midrule
     Learning rate $\eta$  & 0.01 & 0.01 & 0.1 \\
     Dropout rate of MLP layers  & 0.5 & 0.5 & 0.8 \\
     Dropout rate of encoder & 0.6 & 0.6 & 0.5 \\
     Dropout rate of Path Weight Adjacency matrix & 0.6 & 0.6 & 0.6 \\
     Dropout rate of Adjacency matrix & 0.5 & 0.5 & 0.5 \\
     LSTM hidden units & 128 & 128 & 128 \\
     Early stop epochs & 300  & 200 & 100 \\
     L2 weight decay rate & 5e-4  & 5e-4 & 5e-4 \\
     Triplet loss coefficient $\lambda_1$ in Eq.~\eqref{eqUnsupLoss} & 1.0 & 1.0 & 1.0 \\
     Consistency loss coefficient in Eq.~\eqref{eqUnsupLoss} & 1.0  & 1.0 & 1.0 \\
     Tradeoff $\lambda$, $\dot{\mathbf{F}} \oplus \lambda \hat{\mathbf{F}}$ & 1.0 & 10.0 & 1.0 \\
     Embedding dimension (output dimension of Encoder) & 512 & 512 & 256 \\
     \midrule
     Batch size of path sampling & 300 & 300 & 500 \\
     Path length of path sampling  & 12 & 10 & 6 \\
     Window size of path sampling & 8 & 5 & 5 \\
     Orders/Layers of Multi-hop GNN module  & 8 & 4 & 5 \\
     \midrule
     Number of negative sampling for triplet loss & 5000 & 10000 & 5000 \\
     Number of positive sampling for triplet loss & 15000 & 10000 & 5000 \\
     Margin of negative samples for triplet loss  & 0.1 & 1 & 1 \\
     \midrule
     Temperature of consistency loss $T$ & 0.5 & 0.5 & 0.2 \\
     Regularization times $S$ & 4 & 4 & 4 \\
    \bottomrule
    \end{tabular}
\end{table}

\bibliographystyle{cas-model2-names}

\bibliography{cas-refs}

\begin{thebibliography}{50}
\expandafter\ifx\csname natexlab\endcsname\relax\def\natexlab#1{#1}\fi
\providecommand{\url}[1]{\texttt{#1}}
\providecommand{\href}[2]{#2}
\providecommand{\path}[1]{#1}
\providecommand{\DOIprefix}{doi:}
\providecommand{\ArXivprefix}{arXiv:}
\providecommand{\URLprefix}{URL: }
\providecommand{\Pubmedprefix}{pmid:}
\providecommand{\doi}[1]{\href{http://dx.doi.org/#1}{\path{#1}}}
\providecommand{\Pubmed}[1]{\href{pmid:#1}{\path{#1}}}
\providecommand{\bibinfo}[2]{#2}
\ifx\xfnm\relax \def\xfnm[#1]{\unskip,\space#1}\fi
\bibitem[{Berthelot et~al.(2019)Berthelot, Carlini, Goodfellow, Papernot,
  Oliver and Raffel}]{berthelot2019mixmatch}
\bibinfo{author}{Berthelot, D.}, \bibinfo{author}{Carlini, N.},
  \bibinfo{author}{Goodfellow, I.}, \bibinfo{author}{Papernot, N.},
  \bibinfo{author}{Oliver, A.}, \bibinfo{author}{Raffel, C.},
  \bibinfo{year}{2019}.
\newblock \bibinfo{title}{Mixmatch: A holistic approach to semi-supervised
  learning}, in: \bibinfo{booktitle}{NIPS}, pp. \bibinfo{pages}{5050--5060}.
\bibitem[{Chen et~al.(2020a)Chen, Lin, Li, Li, Zhou and Sun}]{2019Measuring}
\bibinfo{author}{Chen, D.}, \bibinfo{author}{Lin, Y.}, \bibinfo{author}{Li,
  W.}, \bibinfo{author}{Li, P.}, \bibinfo{author}{Zhou, J.},
  \bibinfo{author}{Sun, X.}, \bibinfo{year}{2020}a.
\newblock \bibinfo{title}{Measuring and relieving the over-smoothing problem
  for graph neural networks from the topological view}, in:
  \bibinfo{booktitle}{AAAI}, pp. \bibinfo{pages}{3438--3445}.
\bibitem[{Chen et~al.(2022)Chen, Ling, Xu, Ren, Huang, Pu, Hao, Yu and
  He}]{chen2022variational}
\bibinfo{author}{Chen, J.}, \bibinfo{author}{Ling, Y.}, \bibinfo{author}{Xu,
  J.}, \bibinfo{author}{Ren, Y.}, \bibinfo{author}{Huang, S.},
  \bibinfo{author}{Pu, X.}, \bibinfo{author}{Hao, Z.}, \bibinfo{author}{Yu,
  P.S.}, \bibinfo{author}{He, L.}, \bibinfo{year}{2022}.
\newblock \bibinfo{title}{Variational graph generator for multi-view graph
  clustering}.
\newblock \bibinfo{journal}{arXiv preprint arXiv:2210.07011} .
\bibitem[{Chen et~al.(2018)Chen, Ma and Xiao}]{Chen2018fastGCN}
\bibinfo{author}{Chen, J.}, \bibinfo{author}{Ma, T.}, \bibinfo{author}{Xiao,
  C.}, \bibinfo{year}{2018}.
\newblock \bibinfo{title}{Fastgcn: Fast learning with graph convolutional
  networks via importance sampling}, in: \bibinfo{booktitle}{ICLR}.
\bibitem[{Chen et~al.(2023)Chen, Yang, Pu, Ren, Pu, Gao and
  He}]{chen2023shared}
\bibinfo{author}{Chen, J.}, \bibinfo{author}{Yang, Z.}, \bibinfo{author}{Pu,
  J.}, \bibinfo{author}{Ren, Y.}, \bibinfo{author}{Pu, X.},
  \bibinfo{author}{Gao, L.}, \bibinfo{author}{He, L.}, \bibinfo{year}{2023}.
\newblock \bibinfo{title}{Shared-attribute multi-graph clustering with global
  self-attention}, in: \bibinfo{booktitle}{International Conference on Neural
  Information Processing, Part I (ICONIP)}, pp. \bibinfo{pages}{51--63}.
\bibitem[{Chen et~al.(2020b)Chen, Wei, Huang, Ding and Li}]{chen2020simple}
\bibinfo{author}{Chen, M.}, \bibinfo{author}{Wei, Z.}, \bibinfo{author}{Huang,
  Z.}, \bibinfo{author}{Ding, B.}, \bibinfo{author}{Li, Y.},
  \bibinfo{year}{2020}b.
\newblock \bibinfo{title}{Simple and deep graph convolutional networks}, in:
  \bibinfo{booktitle}{ICML}, pp. \bibinfo{pages}{1725--1735}.
\bibitem[{Ding et~al.(2018)Ding, Tang and Zhang}]{2018Ding}
\bibinfo{author}{Ding, M.}, \bibinfo{author}{Tang, J.}, \bibinfo{author}{Zhang,
  J.}, \bibinfo{year}{2018}.
\newblock \bibinfo{title}{Semi-supervised learning on graphs with generative
  adversarial nets}, in: \bibinfo{booktitle}{CIKM}, pp.
  \bibinfo{pages}{913--922}.
\bibitem[{Feng et~al.(2020)Feng, Zhang, Dong, Han, Luan, Xu, Yang, Kharlamov
  and Tang}]{NEURIPS2020_fb4c835f}
\bibinfo{author}{Feng, W.}, \bibinfo{author}{Zhang, J.}, \bibinfo{author}{Dong,
  Y.}, \bibinfo{author}{Han, Y.}, \bibinfo{author}{Luan, H.},
  \bibinfo{author}{Xu, Q.}, \bibinfo{author}{Yang, Q.},
  \bibinfo{author}{Kharlamov, E.}, \bibinfo{author}{Tang, J.},
  \bibinfo{year}{2020}.
\newblock \bibinfo{title}{Graph random neural networks for semi-supervised
  learning on graphs}, in: \bibinfo{booktitle}{NeurIPS}.
\bibitem[{Gao et~al.(2019)Gao, Yang, Zhang, Zhou and
  Hu}]{DBLP:journals/corr/abs-1904-09981}
\bibinfo{author}{Gao, Y.}, \bibinfo{author}{Yang, H.}, \bibinfo{author}{Zhang,
  P.}, \bibinfo{author}{Zhou, C.}, \bibinfo{author}{Hu, Y.},
  \bibinfo{year}{2019}.
\newblock \bibinfo{title}{Graphnas: Graph neural architecture search with
  reinforcement learning}.
\newblock \bibinfo{journal}{CoRR} \bibinfo{volume}{abs/1904.09981}.
\bibitem[{Geng et~al.(2022)Geng, He, Xu and Yu}]{GENG2022126}
\bibinfo{author}{Geng, X.}, \bibinfo{author}{He, X.}, \bibinfo{author}{Xu, L.},
  \bibinfo{author}{Yu, J.}, \bibinfo{year}{2022}.
\newblock \bibinfo{title}{Graph correlated attention recurrent neural network
  for multivariate time series forecasting}.
\newblock \bibinfo{journal}{Information Sciences} \bibinfo{volume}{606},
  \bibinfo{pages}{126--142}.
\bibitem[{Gilmer et~al.(2017)Gilmer, Schoenholz, Riley, Vinyals and
  Dahl}]{2017neural}
\bibinfo{author}{Gilmer, J.}, \bibinfo{author}{Schoenholz, S.S.},
  \bibinfo{author}{Riley, P.F.}, \bibinfo{author}{Vinyals, O.},
  \bibinfo{author}{Dahl, G.E.}, \bibinfo{year}{2017}.
\newblock \bibinfo{title}{Neural message passing for quantum chemistry}, in:
  \bibinfo{booktitle}{ICML}, pp. \bibinfo{pages}{1263--1272}.
\bibitem[{Grover and Leskovec(2016)}]{2016node2vec}
\bibinfo{author}{Grover, A.}, \bibinfo{author}{Leskovec, J.},
  \bibinfo{year}{2016}.
\newblock \bibinfo{title}{node2vec: Scalable feature learning for networks},
  in: \bibinfo{booktitle}{KDD}, pp. \bibinfo{pages}{855--864}.
\bibitem[{Hamilton et~al.(2017)Hamilton, Ying and
  Leskovec}]{hamilton2017inductive}
\bibinfo{author}{Hamilton, W.L.}, \bibinfo{author}{Ying, Z.},
  \bibinfo{author}{Leskovec, J.}, \bibinfo{year}{2017}.
\newblock \bibinfo{title}{Inductive representation learning on large graphs},
  in: \bibinfo{booktitle}{NIPS}, pp. \bibinfo{pages}{1024--1034}.
\bibitem[{Hao et~al.(2020)Hao, Lu, Huang, Wang, Hu, Liu, Chen and
  Lee}]{202ASGN}
\bibinfo{author}{Hao, Z.}, \bibinfo{author}{Lu, C.}, \bibinfo{author}{Huang,
  Z.}, \bibinfo{author}{Wang, H.}, \bibinfo{author}{Hu, Z.},
  \bibinfo{author}{Liu, Q.}, \bibinfo{author}{Chen, E.}, \bibinfo{author}{Lee,
  C.}, \bibinfo{year}{2020}.
\newblock \bibinfo{title}{Asgn: An active semi-supervised graph neural network
  for molecular property prediction}, in: \bibinfo{booktitle}{KDD}, pp.
  \bibinfo{pages}{731--752}.
\bibitem[{Hassani and Khasahmadi(2020)}]{MVGRL}
\bibinfo{author}{Hassani, K.}, \bibinfo{author}{Khasahmadi, A.H.},
  \bibinfo{year}{2020}.
\newblock \bibinfo{title}{Contrastive multi-view representation learning on
  graphs}, in: \bibinfo{booktitle}{Proceedings of the 37th International
  Conference on Machine Learning}, \bibinfo{publisher}{JMLR.org}.
\bibitem[{He et~al.(2023)He, Bai, Yang, Du and Liang}]{HE2023222}
\bibinfo{author}{He, L.}, \bibinfo{author}{Bai, L.}, \bibinfo{author}{Yang,
  X.}, \bibinfo{author}{Du, H.}, \bibinfo{author}{Liang, J.},
  \bibinfo{year}{2023}.
\newblock \bibinfo{title}{High-order graph attention network}.
\newblock \bibinfo{journal}{Information Sciences} \bibinfo{volume}{630},
  \bibinfo{pages}{222--234}.
\newblock \DOIprefix\doi{https://doi.org/10.1016/j.ins.2023.02.054}.
\bibitem[{He et~al.(2021)He, Ong and Bai}]{he2021learning}
\bibinfo{author}{He, T.}, \bibinfo{author}{Ong, Y.S.}, \bibinfo{author}{Bai,
  L.}, \bibinfo{year}{2021}.
\newblock \bibinfo{title}{Learning conjoint attentions for graph neural nets}.
\bibitem[{He et~al.(2022)He, Zhou, Ong and Cong}]{he2022neighbors}
\bibinfo{author}{He, T.}, \bibinfo{author}{Zhou, H.}, \bibinfo{author}{Ong,
  Y.S.}, \bibinfo{author}{Cong, G.}, \bibinfo{year}{2022}.
\newblock \bibinfo{title}{Not all neighbors are worth attending to: Graph
  selective attention networks for semi-supervised learning}.
\newblock \href{http://arxiv.org/abs/2210.07715}{\tt arXiv:2210.07715}.
\bibitem[{Hochreiter and Schmidhuber(1997)}]{LSTM}
\bibinfo{author}{Hochreiter, S.}, \bibinfo{author}{Schmidhuber, J.},
  \bibinfo{year}{1997}.
\newblock \bibinfo{title}{Long short-term memory}.
\newblock \bibinfo{journal}{Neural Computation} \bibinfo{volume}{9},
  \bibinfo{pages}{1735--1780}.
\bibitem[{Huang et~al.(2022)Huang, Yi, Wang, Li, Peng and Xiong}]{HUANG2022286}
\bibinfo{author}{Huang, F.}, \bibinfo{author}{Yi, P.}, \bibinfo{author}{Wang,
  J.}, \bibinfo{author}{Li, M.}, \bibinfo{author}{Peng, J.},
  \bibinfo{author}{Xiong, X.}, \bibinfo{year}{2022}.
\newblock \bibinfo{title}{A dynamical spatial-temporal graph neural network for
  traffic demand prediction}.
\newblock \bibinfo{journal}{Information Sciences} \bibinfo{volume}{594},
  \bibinfo{pages}{286--304}.
\bibitem[{Juan et~al.(2023)Juan, Zhou, Wang, Jin, Tang and
  Wang}]{JUAN2023118935}
\bibinfo{author}{Juan, X.}, \bibinfo{author}{Zhou, F.}, \bibinfo{author}{Wang,
  W.}, \bibinfo{author}{Jin, W.}, \bibinfo{author}{Tang, J.},
  \bibinfo{author}{Wang, X.}, \bibinfo{year}{2023}.
\newblock \bibinfo{title}{Ins-gnn: Improving graph imbalance learning with
  self-supervision}.
\newblock \bibinfo{journal}{Information Sciences} \bibinfo{volume}{637},
  \bibinfo{pages}{118935}.
\newblock \URLprefix
  \url{https://www.sciencedirect.com/science/article/pii/S0020025523005042},
  \DOIprefix\doi{https://doi.org/10.1016/j.ins.2023.118935}.
\bibitem[{Kim and Oh(2021)}]{2021superGAT}
\bibinfo{author}{Kim, D.}, \bibinfo{author}{Oh, A.}, \bibinfo{year}{2021}.
\newblock \bibinfo{title}{How to find your friendly neighborhood: Graph
  attention design with self-supervision}, in: \bibinfo{booktitle}{ICLR}.
\bibitem[{Kipf and Welling(2017)}]{GCN}
\bibinfo{author}{Kipf, T.N.}, \bibinfo{author}{Welling, M.},
  \bibinfo{year}{2017}.
\newblock \bibinfo{title}{Semi-supervised classification with graph
  convolutional networks}, in: \bibinfo{booktitle}{ICLR}.
\bibitem[{Klicpera et~al.(2019)Klicpera, Bojchevski and
  G{\"{u}}nnemann}]{DBLP:conf/iclr/KlicperaBG19}
\bibinfo{author}{Klicpera, J.}, \bibinfo{author}{Bojchevski, A.},
  \bibinfo{author}{G{\"{u}}nnemann, S.}, \bibinfo{year}{2019}.
\newblock \bibinfo{title}{Predict then propagate: Graph neural networks meet
  personalized pagerank}, in: \bibinfo{booktitle}{ICLR}.
\bibitem[{Lee(2013)}]{2013pesudo}
\bibinfo{author}{Lee, D.H.}, \bibinfo{year}{2013}.
\newblock \bibinfo{title}{Pseudo-label: The simple and efficient
  semi-supervised learning method for deep neural networks}, in:
  \bibinfo{booktitle}{ICML}.
\bibitem[{Li et~al.(2022)Li, Cao, Zhu, Liu, Zhu and Wu}]{LI202250}
\bibinfo{author}{Li, H.}, \bibinfo{author}{Cao, J.}, \bibinfo{author}{Zhu, J.},
  \bibinfo{author}{Liu, Y.}, \bibinfo{author}{Zhu, Q.}, \bibinfo{author}{Wu,
  G.}, \bibinfo{year}{2022}.
\newblock \bibinfo{title}{Curvature graph neural network}.
\newblock \bibinfo{journal}{Information Sciences} \bibinfo{volume}{592},
  \bibinfo{pages}{50--66}.
\bibitem[{Luo et~al.(2024)Luo, Zhao, Qin, Ju and Zhang}]{10138449}
\bibinfo{author}{Luo, X.}, \bibinfo{author}{Zhao, Y.}, \bibinfo{author}{Qin,
  Y.}, \bibinfo{author}{Ju, W.}, \bibinfo{author}{Zhang, M.},
  \bibinfo{year}{2024}.
\newblock \bibinfo{title}{Towards semi-supervised universal graph
  classification}.
\newblock \bibinfo{journal}{IEEE Transactions on Knowledge and Data
  Engineering} \bibinfo{volume}{36}, \bibinfo{pages}{416--428}.
\newblock \DOIprefix\doi{10.1109/TKDE.2023.3280859}.
\bibitem[{Mcpherson et~al.(2001)Mcpherson, Smith-Lovin and
  Cook}]{2001Homohoily}
\bibinfo{author}{Mcpherson, M.}, \bibinfo{author}{Smith-Lovin, L.},
  \bibinfo{author}{Cook, J.M.}, \bibinfo{year}{2001}.
\newblock \bibinfo{title}{Birds of a feather: Homophily in social networks}.
\newblock \bibinfo{journal}{Annual Review of Sociology} \bibinfo{volume}{27},
  \bibinfo{pages}{415--444}.
\bibitem[{NT and Maehara(2019)}]{nt2019revisiting}
\bibinfo{author}{NT, H.}, \bibinfo{author}{Maehara, T.}, \bibinfo{year}{2019}.
\newblock \bibinfo{title}{Revisiting graph neural networks: All we have is
  low-pass filters}.
\newblock \bibinfo{journal}{CoRR} \bibinfo{volume}{abs/1905.09550}.
\bibitem[{Olivier et~al.(2009)Olivier, Bernhard and Alexander}]{2009Semi}
\bibinfo{author}{Olivier, C.}, \bibinfo{author}{Bernhard, S.},
  \bibinfo{author}{Alexander, Z.}, \bibinfo{year}{2009}.
\newblock \bibinfo{title}{Semi-supervised learning}.
\newblock \bibinfo{journal}{Journal of the Royal Statistical Society}
  \bibinfo{volume}{172}, \bibinfo{pages}{530--530}.
\bibitem[{Oono and Suzuki(2020)}]{oono2021graph}
\bibinfo{author}{Oono, K.}, \bibinfo{author}{Suzuki, T.}, \bibinfo{year}{2020}.
\newblock \bibinfo{title}{Graph neural networks exponentially lose expressive
  power for node classification}, in: \bibinfo{booktitle}{ICLR}.
\bibitem[{Qu et~al.(2019)Qu, Bengio and Tang}]{DBLP:conf/icml/QuBT19}
\bibinfo{author}{Qu, M.}, \bibinfo{author}{Bengio, Y.}, \bibinfo{author}{Tang,
  J.}, \bibinfo{year}{2019}.
\newblock \bibinfo{title}{{GMNN:} graph markov neural networks}, in:
  \bibinfo{booktitle}{ICML}, pp. \bibinfo{pages}{5241--5250}.
\bibitem[{Rong et~al.(2020)Rong, Huang, Xu and Huang}]{rong2020dropedge}
\bibinfo{author}{Rong, Y.}, \bibinfo{author}{Huang, W.}, \bibinfo{author}{Xu,
  T.}, \bibinfo{author}{Huang, J.}, \bibinfo{year}{2020}.
\newblock \bibinfo{title}{Dropedge: Towards deep graph convolutional networks
  on node classification}, in: \bibinfo{booktitle}{ICLR}.
\bibitem[{Schroff et~al.(2015)Schroff, Kalenichenko and Philbin}]{2015tirloss}
\bibinfo{author}{Schroff, F.}, \bibinfo{author}{Kalenichenko, D.},
  \bibinfo{author}{Philbin, J.}, \bibinfo{year}{2015}.
\newblock \bibinfo{title}{Facenet: A unified embedding for face recognition and
  clustering}, in: \bibinfo{booktitle}{CVPR}, pp. \bibinfo{pages}{815--823}.
\bibitem[{Shchur et~al.(2018)Shchur, Mumme, Bojchevski and
  G{\"{u}}nnemann}]{2018Pitfalls}
\bibinfo{author}{Shchur, O.}, \bibinfo{author}{Mumme, M.},
  \bibinfo{author}{Bojchevski, A.}, \bibinfo{author}{G{\"{u}}nnemann, S.},
  \bibinfo{year}{2018}.
\newblock \bibinfo{title}{Pitfalls of graph neural network evaluation}.
\newblock \bibinfo{journal}{CoRR} \bibinfo{volume}{abs/1811.05868}.
\bibitem[{Sun et~al.(2019)Sun, Hoffman, Verma and Tang}]{InfoGraph}
\bibinfo{author}{Sun, F.Y.}, \bibinfo{author}{Hoffman, J.},
  \bibinfo{author}{Verma, V.}, \bibinfo{author}{Tang, J.},
  \bibinfo{year}{2019}.
\newblock \bibinfo{title}{Infograph: Unsupervised and semi-supervised
  graph-level representation learning via mutual information maximization}, in:
  \bibinfo{booktitle}{International Conference on Learning Representations}.
\bibitem[{Veličković et~al.(2018)Veličković, Cucurull, Casanova, Romero,
  Liò and Bengio}]{GAT}
\bibinfo{author}{Veličković, P.}, \bibinfo{author}{Cucurull, G.},
  \bibinfo{author}{Casanova, A.}, \bibinfo{author}{Romero, A.},
  \bibinfo{author}{Liò, P.}, \bibinfo{author}{Bengio, Y.},
  \bibinfo{year}{2018}.
\newblock \bibinfo{title}{Graph attention networks}, in:
  \bibinfo{booktitle}{ICLR}.
\bibitem[{Veličković et~al.(2019)Veličković, Fedus, Hamilton, Liò, Bengio
  and Hjelm}]{DGI}
\bibinfo{author}{Veličković, P.}, \bibinfo{author}{Fedus, W.},
  \bibinfo{author}{Hamilton, W.L.}, \bibinfo{author}{Liò, P.},
  \bibinfo{author}{Bengio, Y.}, \bibinfo{author}{Hjelm, R.D.},
  \bibinfo{year}{2019}.
\newblock \bibinfo{title}{Deep graph infomax}, in:
  \bibinfo{booktitle}{International Conference on Learning Representations}.
\newblock \URLprefix \url{https://openreview.net/forum?id=rklz9iAcKQ}.
\bibitem[{Verma et~al.(2021)Verma, Qu, Kawaguchi, Lamb, Bengio, Kannala and
  Tang}]{2019GraphMix}
\bibinfo{author}{Verma, V.}, \bibinfo{author}{Qu, M.},
  \bibinfo{author}{Kawaguchi, K.}, \bibinfo{author}{Lamb, A.},
  \bibinfo{author}{Bengio, Y.}, \bibinfo{author}{Kannala, J.},
  \bibinfo{author}{Tang, J.}, \bibinfo{year}{2021}.
\newblock \bibinfo{title}{Graphmix: Improved training of gnns for
  semi-supervised learning}, in: \bibinfo{booktitle}{AAAI},
  \bibinfo{publisher}{{AAAI} Press}. pp. \bibinfo{pages}{10024--10032}.
\bibitem[{Wei et~al.(2023)Wei, Wang, Fu, Hu and Li}]{WEI2023166}
\bibinfo{author}{Wei, Q.}, \bibinfo{author}{Wang, J.}, \bibinfo{author}{Fu,
  X.}, \bibinfo{author}{Hu, J.}, \bibinfo{author}{Li, X.},
  \bibinfo{year}{2023}.
\newblock \bibinfo{title}{Aic-gnn: Adversarial information completion for graph
  neural networks}.
\newblock \bibinfo{journal}{Information Sciences} \bibinfo{volume}{626},
  \bibinfo{pages}{166--179}.
\newblock \URLprefix
  \url{https://www.sciencedirect.com/science/article/pii/S0020025522016073},
  \DOIprefix\doi{https://doi.org/10.1016/j.ins.2022.12.112}.
\bibitem[{Wu et~al.(2019)Wu, Jr., Zhang, Fifty, Yu and
  Weinberger}]{DBLP:conf/icml/WuSZFYW19}
\bibinfo{author}{Wu, F.}, \bibinfo{author}{Jr., A.H.S.},
  \bibinfo{author}{Zhang, T.}, \bibinfo{author}{Fifty, C.},
  \bibinfo{author}{Yu, T.}, \bibinfo{author}{Weinberger, K.Q.},
  \bibinfo{year}{2019}.
\newblock \bibinfo{title}{Simplifying graph convolutional networks}, in:
  \bibinfo{booktitle}{ICML}, pp. \bibinfo{pages}{6861--6871}.
\bibitem[{Xu et~al.(2021)Xu, Wang, Ni, Zhang and Tang}]{xu2021graphsad}
\bibinfo{author}{Xu, M.}, \bibinfo{author}{Wang, H.}, \bibinfo{author}{Ni, B.},
  \bibinfo{author}{Zhang, W.}, \bibinfo{author}{Tang, J.},
  \bibinfo{year}{2021}.
\newblock \bibinfo{title}{Graphsad: Learning graph representations with
  structure-attribute disentanglement}, in: \bibinfo{booktitle}{ICLR}.
\bibitem[{Yang et~al.(2021)Yang, Ma and Cheng}]{yang2020rethinking}
\bibinfo{author}{Yang, H.}, \bibinfo{author}{Ma, K.}, \bibinfo{author}{Cheng,
  J.}, \bibinfo{year}{2021}.
\newblock \bibinfo{title}{Rethinking graph regularization for graph neural
  networks}, in: \bibinfo{booktitle}{AAAI}, \bibinfo{publisher}{{AAAI} Press}.
  pp. \bibinfo{pages}{4573--4581}.
\bibitem[{Yang et~al.(2016)Yang, Cohen and Salakhutdinov}]{2016Revisiting}
\bibinfo{author}{Yang, Z.}, \bibinfo{author}{Cohen, W.W.},
  \bibinfo{author}{Salakhutdinov, R.}, \bibinfo{year}{2016}.
\newblock \bibinfo{title}{Revisiting semi-supervised learning with graph
  embeddings}, in: \bibinfo{booktitle}{ICML}, pp. \bibinfo{pages}{40--48}.
\bibitem[{Zhang et~al.(2018)Zhang, Cisse, Dauphin and
  Lopez-Paz}]{zhang2018mixup}
\bibinfo{author}{Zhang, H.}, \bibinfo{author}{Cisse, M.},
  \bibinfo{author}{Dauphin, Y.N.}, \bibinfo{author}{Lopez-Paz, D.},
  \bibinfo{year}{2018}.
\newblock \bibinfo{title}{mixup: Beyond empirical risk minimization}, in:
  \bibinfo{booktitle}{ICLR}.
\bibitem[{Zhang et~al.(2020)Zhang, Zhu, Wang and Zhang}]{Zhang2020Adaptive}
\bibinfo{author}{Zhang, K.}, \bibinfo{author}{Zhu, Y.}, \bibinfo{author}{Wang,
  J.}, \bibinfo{author}{Zhang, J.}, \bibinfo{year}{2020}.
\newblock \bibinfo{title}{Adaptive structural fingerprints for graph attention
  networks}, in: \bibinfo{booktitle}{ICLR}.
\bibitem[{Zhou et~al.(2023)Zhou, Gong, Wang, Gao and Zhao}]{SMGCL}
\bibinfo{author}{Zhou, H.}, \bibinfo{author}{Gong, M.}, \bibinfo{author}{Wang,
  S.}, \bibinfo{author}{Gao, Y.}, \bibinfo{author}{Zhao, Z.},
  \bibinfo{year}{2023}.
\newblock \bibinfo{title}{Smgcl: Semi-supervised multi-view graph contrastive
  learning}.
\newblock \bibinfo{journal}{Knowledge-Based Systems} \bibinfo{volume}{260},
  \bibinfo{pages}{110120}.
\bibitem[{Zhu et~al.(2019)Zhu, Zhang, Cui and Zhu}]{2019attckGCN}
\bibinfo{author}{Zhu, D.}, \bibinfo{author}{Zhang, Z.}, \bibinfo{author}{Cui,
  P.}, \bibinfo{author}{Zhu, W.}, \bibinfo{year}{2019}.
\newblock \bibinfo{title}{Robust graph convolutional networks against
  adversarial attacks}, in: \bibinfo{booktitle}{KDD}, p.
  \bibinfo{pages}{1399–1407}.
\bibitem[{Zhu and Koniusz(2021)}]{zhu2021simple}
\bibinfo{author}{Zhu, H.}, \bibinfo{author}{Koniusz, P.}, \bibinfo{year}{2021}.
\newblock \bibinfo{title}{Simple spectral graph convolution}, in:
  \bibinfo{booktitle}{ICLR}.
\bibitem[{Z{\"{u}}gner et~al.(2018)Z{\"{u}}gner, Akbarnejad and
  G{\"{u}}nnemann}]{2018attack}
\bibinfo{author}{Z{\"{u}}gner, D.}, \bibinfo{author}{Akbarnejad, A.},
  \bibinfo{author}{G{\"{u}}nnemann, S.}, \bibinfo{year}{2018}.
\newblock \bibinfo{title}{Adversarial attacks on neural networks for graph
  data}, in: \bibinfo{booktitle}{KDD}, p. \bibinfo{pages}{2847–2856}.

\end{thebibliography}

\end{document}